\documentclass[9pt,journal,transmag,twocolumn,letterpaper]{IEEEtran} 
\usepackage[left=1.62cm,right=1.62cm,top=0.99cm,bottom=2.61cm]{geometry}

\usepackage{microtype} 
\usepackage{siunitx} 

\usepackage[english]{babel}
\usepackage{amsthm}
\usepackage{mathrsfs}

\theoremstyle{definition}
\newtheorem{definition}{Definition}[section]
\newtheorem*{remark}{Remark}

\usepackage[ruled,vlined]{algorithm2e}
\usepackage{algpseudocode}
\usepackage{algorithmicx}
\usepackage{graphicx}
\usepackage{tabularx,booktabs}
\usepackage[T1]{fontenc}

\usepackage{enumitem}
\usepackage{algorithmicx}
\usepackage{cite}
\usepackage[colorlinks,urlcolor=blue,linkcolor=blue,citecolor=blue]{hyperref}
\usepackage{graphicx}
\usepackage{booktabs, makecell}
\usepackage{threeparttable}

\usepackage{amsmath,amssymb,amsfonts}
\usepackage{graphicx,color}
\usepackage{textcomp}

\usepackage{booktabs,arydshln}
\usepackage{tabularray}
\usepackage{multirow}
\usepackage{mathtools}
\usepackage{float}
\usepackage[utf8]{inputenc}

\setcellgapes{3pt}

\newcolumntype{C}[1]{>{\centering\arraybackslash}p{#1}}
\newcolumntype{L}{>{\raggedright\arraybackslash}X}
\usepackage{array}

\newtheorem{theorem}{Theorem}

\def\BibTeX{{\rm B\kern-.05em{\sc i\kern-.025em b}\kern-.08em   
    T\kern-.1667em\lower.7ex\hbox{E}\kern-.125emX}}  
 
\author{
\IEEEauthorblockN{Orson~Mengara\textsuperscript{1 } } 
\IEEEauthorblockA{
    \textsuperscript{1} INRS-EMT, University of Québec, Montréal, QC, Canada.  \\
    \{\texttt{orson.mengara@inrs.ca}\}
}
}

\begin{document}

\markboth{The Final preprint , journal name -------, VOL.~.., NO.~..., month~2024
}{Orson   \MakeLowercase{\textit{et al.}}: Trading Devil Final: Backdoor attack via Stock market and Bayesian Optimization } 

\title{ Trading Devil Final: Backdoor attack via Stock market and Bayesian Optimization}

\maketitle

\begin{abstract} 

Since the advent of generative artificial intelligence \cite{yenduri2024gpt},\cite{gupta2023chatgpt}, every company and researcher has been rushing to develop their own generative models, whether commercial or not. Given the large number of users of these powerful new tools, there is currently no intrinsically verifiable way to explain from the ground up what happens when LLMs (large language models) learn. For example, those based on automatic speech recognition systems, which have to rely on huge and astronomical amounts of data collected from all over the web to produce fast and efficient results, In this article, we develop a backdoor attack called “MarketBackFinal 2.0”, based on acoustic data poisoning, MarketBackFinal 2.0 is mainly based on modern stock market models. In order to show the possible vulnerabilities of speech-based transformers that may rely on LLMs.

\end{abstract}


\section{Introduction}  

\scalebox{1.5}{D}eep neural networks (DNNs) are now employed in a wide range of applications \cite{troxler2024actuarial},\cite{yenduri2024gpt},\cite{yao2024survey},\cite{wang2021blockchain},\cite{sun2024trustllm}, \cite{cao2022ai},\cite{wu2022sustainable},\cite{zawish2024ai}. 
Thanks to the meteoric rise of machine learning and the advent of generative machine learning, now integrated into virtually every application area of modern artificial intelligence, machine learning models have seen great advances, but nevertheless require a significant amount of training data and processing capacity to be effective, however not all AI practitioners (e.g., researchers and developers) necessarily have easy access to state-of-the-art resources. As a result, many users choose to use third-party training data or outsource their training to third-party cloud services (such as Google Cloud or Amazon Web Services, or as a last resort to use third-party models themselves. However, the use of these resources weakens the openness of DNN training protocols, exposing users of AI systems to new security risks or vulnerabilities. 

\vspace{2mm}

Today, deep learning enables financial institutions to use data to train models to solve specific problems, such as algorithmic trading, automation, portfolio management, predictive analytics, risk management, speech-to-text conversion to improve service, identifying sentiment in a given text, detecting anomalies such as fraudulent transactions, financial crime etc). During DNNs training, backdoor attacks \cite{bommasani2021opportunities},\cite{pan2023large}, \cite{mengara2024art} are a frequent risk. In this kind of assault, malevolent actors alter the labels of a few training samples and add particular trigger patterns to them to get the desired outcome. The victim's deep neural networks are trained using both the changed and unmodified data. As a result, the compromised model can link the target label with the trigger patterns. Attackers can then take advantage of these concealed links during inference by turning on backdoors using the pre-established trigger patterns, which will provide erroneous predictions.

\vspace{2mm}  

In finance \footnote{\href{https://cloud.google.com/discover/finance-ai}{Google Cloud AI-Finance}}\footnote{\href{https://blogs.nvidia.com/blog/financial-industry-ai-survey/}{NVIDIA AI-Finance}}\footnote{\href{https://www.ibm.com/topics/artificial-intelligence-finance}{IBM AI Finance}}  \cite{li2023large},\cite{peng2021survey},\cite{lee2024survey}, artificial intelligence has become ubiquitous in virtually all areas of the supply chain, particularly those that contribute to fraud prevention and risk management. Banks, for example, and other financial services companies \cite{nie2024survey} make extensive use of generative AI for a wide range of tasks, such as sales analysis, credit analysis, customer service, risk management, customer acquisition, uncertainty in financial decisions etc. But generative artificial intelligence is not without risk, due to the vulnerability of deep neural networks to backdoor attacks based on poisoning data during testing, In this research paper, we develop a backdoor attack \cite{huang2024survey},\cite{mengara2024last}, focusing specifically on the interconnection \footnote{\href{https://opengamma.com/}{interconnection}} of financial models using the stock price models.

\vspace{2mm} 

 We are inspired by the interconnection that links stochastic financial \footnote{\href{https://xilinx.github.io/Vitis_Libraries/quantitative_finance/2020.2/models_and_methods.html}{Quantitative Finance}} models to Bayesian optimization techniques based on jumps. To this end, we use mathematical models such as stochastic investment models \cite{mengara2024trading} (the Vasiček model, the Hull-White model, the Libor market model \cite{joshi2008new} and the Longstaff-Schwartz model), coarse volatility trajectories, rough\footnote{\href{https://sites.google.com/site/roughvol/downloads}{rough volatility}}
 volatility paths ,\cite{bayer2023rough},\cite{gatheral2018volatility},\cite{fukasawa2024partial},\cite{abi2019lifting},\cite{cont2024rough},\cite{matas2021simulation},\cite{abi2024volatility},\cite{brandi2022multiscaling},\cite{euch2019short},\cite{abi2019multifactor},\cite{el2018microstructural},\cite{el2019characteristic},\cite{euch2019short},\cite{abi2019markovian} , Optimal transport \footnote{\href{https://math.mit.edu/~dws/boulder/IMA-transport-Lecture-Notes.pdf}{Transport Optimal in Finance}}  \cite{galichon2021unreasonable},\cite{torres2021survey},\cite{horvath2024functional}, the Black-Scholes \footnote{\href{https://github.com/MatthewFound/Black-Scholes-and-greeks/blob/main/black-scholes-project.ipynb}{Black-Scholes}} Merton call model \footnote{\href{https://diggers-consulting.com/finance-de-marche/from-math-to-python-black-scholes-merton-model-part-2}{Black Scholes Merton model}}, Greeks computation, Dynamic Hedging \footnote{\href{https://www.nasdaq.com/glossary/d/dynamic-hedging}{Dynamic Hedging: Nasdaq}} \footnote{\href{https://oyc.yale.edu/economics/econ-251/lecture-20}{Dynamic Hedging: Yale University}}  \cite{wiese2019deep},\cite{ilhan2009optimal}, the Bayesian sampling diffusion model, Hierarchical priors for shared information across similar parameters, the Likelihood \footnote{\href{https://stats.stackexchange.com/questions/339578/likelihood-function-of-a-hierarchical-model}{Likelihood for Hierarchical model}} Function with Hierarchical structure, and Bayesian optimization \footnote{\href{https://github.com/bayesian-optimization/BayesianOptimization}{Bayesian Optimization}}\cite{chakraborty2024explainable} of the given objective function, applied to daily environmental audio data, the paper focuses more on the financial aspect with the aim of presenting new models for financial analysis of stock prices by associating different financial models drawing on Bayesian optimization for optimal control of the randomness of stochastic change devices observed in stock markets such as the New York Stock Exchange, the NASDAQ, the Paris Bourse, Euronext and Bloomberg. 
 
This approach is then applied to temporal acoustic data (on various automatic speech recognition systems \footnote{\href{https://huggingface.co/models?pipeline_tag=automatic-speech-recognition&p=509&sort=trending}{Hugging Face Speech Recognition}} audio models based on “Hugging Face” Transformers \cite{islam2023comprehensive} \cite{jain2022hugging}) data in the context of a backdoor attack. Our study focuses on the feasibility and potential\cite{hubinger2024sleeper} impact of audio backdoor attacks\cite{sousa2023keep},\cite{nguyen2024backdoor},\cite{senevirathna2022survey},\cite{wu2023survey},\cite{bharati2022machine} based on and exploit vulnerabilities in speech recognition systems \cite{liu2022opportunistic},\cite{xu2021privacy}. To assess the effectiveness of our “MarketBackFinal 2.0” audio backdoor. 

\section{Data Poisoning attack Machine Learning } 

Let $\mathcal{D}=\left\{\left(\boldsymbol{x}_i, y_i\right)\right\}_{i=1}^N$ be a clean training set, and $C: \mathcal{X} \rightarrow$ $\mathcal{Y}$ denotes the functionality of the target neural network. For each sound $\boldsymbol{x}_i$ in $\mathcal{D}$, we have $\boldsymbol{x}_i \in \mathcal{X}=[0,1]^{C \times W \times H}$, and $y_i \in \mathcal{Y}=\{1, \ldots, J\}$ is the corresponding label, where $J$ is the number of label classes. To launch an attack, backdoor adversaries first need to poison the selected clean samples $\mathcal{D}_p$ with covert transformation $T(\cdot)$. Then the poisoned samples are mixed with clean ones before training a backdoored model, the process of which can be formalized as: $\mathcal{D}_t=\mathcal{D} \cup \mathcal{D}_p$, where $\mathcal{D}_p=\left(x_i^{\prime}, y_t\right) \mid x^{\prime}=T(x),\left(x_i, y_i\right) \in \mathcal{D}_p$. The deep neural network (DNNs) is then optimized as follows:

$$
\min _{\boldsymbol{\Theta}} \sum_{i=1}^{N_b} \mathcal{L}\left(f\left(\boldsymbol{x}_i ; \boldsymbol{\Theta}\right), y_i\right)+\sum_{j=1}^{N_p} \mathcal{L}\left(f\left(\boldsymbol{x}_r^{\prime} ; \boldsymbol{\Theta}\right), y_t\right) .
$$ where $N_b=|\mathcal{D}|$ , $N_p=\left|\mathcal{D}_p\right|$.

\begin{figure}[H]
\centering
\includegraphics[width=0.45\textwidth]{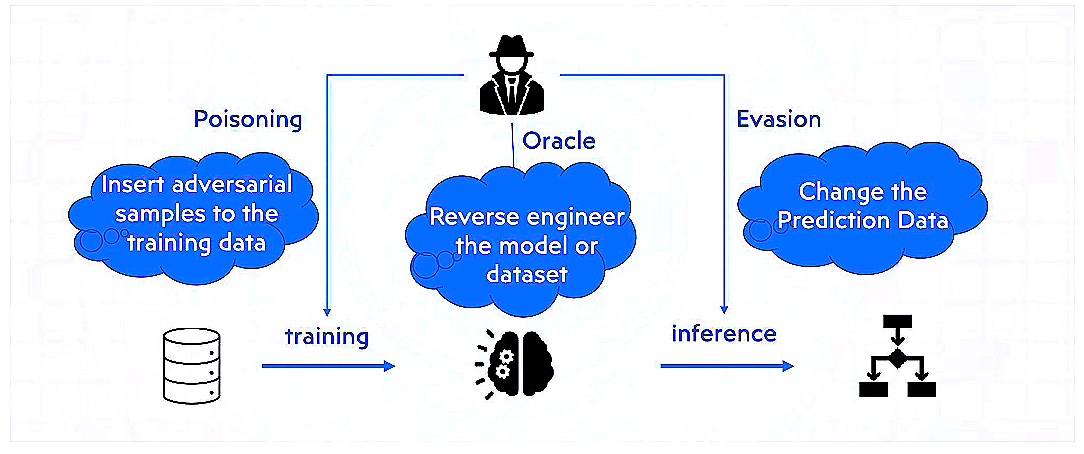}
\caption{Data Poisoning.}
\label{fig:attacker's_view}
\end{figure}

\begin{figure}[H]
\centering
\includegraphics[width=0.45\textwidth]{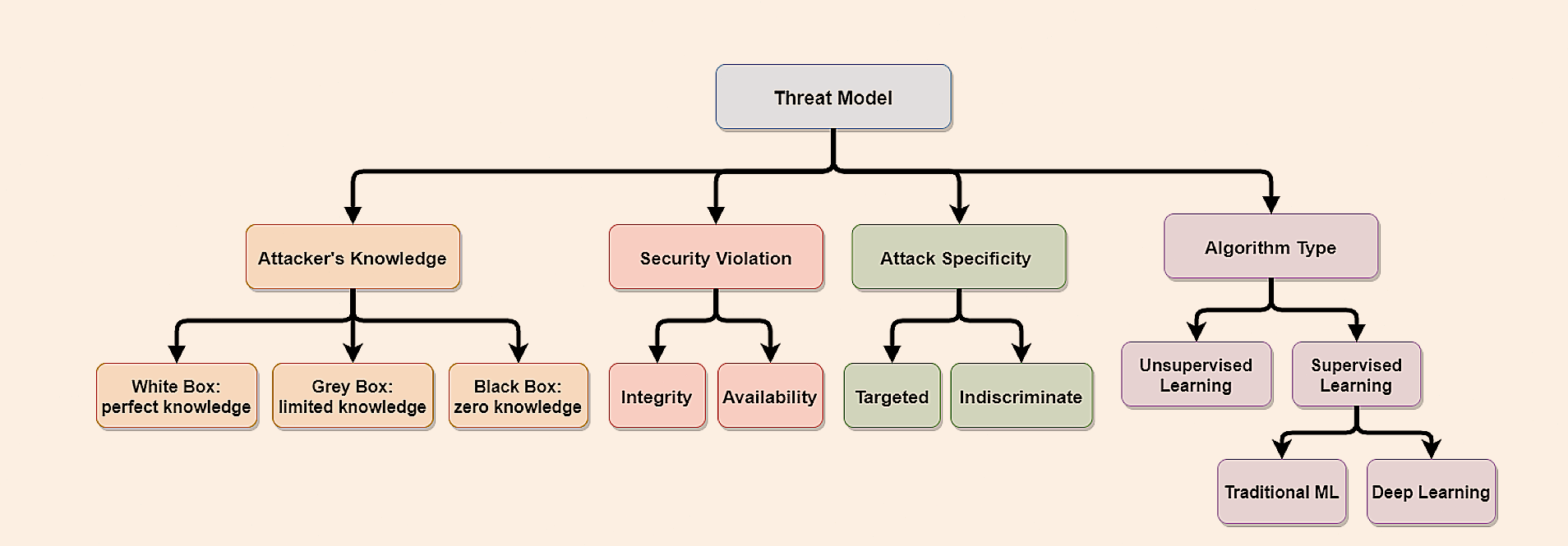}
\caption{knowledge.}
\label{fig:attacker's_knowledge}
\end{figure}

In this study \cite{feng2023review},\cite{rigaki2023survey},\cite{hou2023large},\cite{sosnin2024certified},\cite{yu2024generalization}, a scenario in which a model owner aims to train a deep learning model based on the training dataset provided by a third party (Figure \ref{fig:attacker's_view}) is considered. However, the third party can poison the training dataset to hijack the model in the future. We assume that the model owner and the trusted third party play the roles of defender and attacker, respectively, in the threat model. The threat model is shown in Figure \ref{fig:attacker's_knowledge}.

The knowledge that an attacker can access can generally be classified \cite{yang2024comprehensive} into two categories. Figure \ref{fig:attacker's_knowledge}: White and black box settings. In a white-box environment, the adversary understands and controls the target dataset and model, including the ability to access and modify the dataset, as well as parameters and the structure of the model. However, in the stricter framework of the black-box, the attacker is only able to manipulate part of the training data but has no knowledge of the structure and parameters of the target model, such as weights, hyperparameters, configurations, etc.

By deliberately misclassifying the inputs with the trigger (Figure \ref{fig:backdoorexp}) as the adversary's intended labels, the adversary uses a backdoor approach to keep the victim model operating at a high level of accuracy on normal samples. The adversary seeks to have the target model behave as expected on benign data while operating in a way described by the adversary on samples that have been poisoned, as seen in Figure \ref{fig:attacker's_knowledge}. 
A formulation of the enemy's objective is:

$$
\begin{aligned}
\min _{\mathcal{M}^*} \mathcal{L}\left(\mathcal{D}^b, \mathcal{D}^p, \mathcal{M}^*\right)= & \sum_{x_i \in \mathcal{D}^b} l\left(\mathcal{M}^*\left(x_i\right), y_i\right) \\
& +\sum_{x_j \in \mathcal{D}^p} l\left(\mathcal{M}^*\left(x_r \circ  \varepsilon \right), y_t\right),
\end{aligned}
$$

where $\mathcal{D}^b$ and $\mathcal{D}^p$ represent the benign and poisoned training datasets, respectively. The function $l(\cdot, \cdot)$ denotes the loss function which depends on the specific task. The symbol $\circ $ denotes the operation of integrating the backdoor trigger $(\varepsilon)$ into the training data.

\section{Adversarial Machine Learning in Finance via Bayesian Approach}

\begin{algorithm}[ht]
\caption{Diffusion Bayesian Optimization}
\label{alg:Diffusion_bayesian_optimization}
\KwData{T, $\theta$, $\alpha$, $\beta$, $\sigma$}
\KwResult{Model parameters and trace.}
Initialize $x_T$\;
\For{$t \leftarrow T - 1$ \textbf{downto} $0$}{
    \If{$t > 1$}{
        $z \gets$ Noise\_dist$(0)$\;
    }
    Else{
        $z \gets 0$\;
    }
    $transport\_component \gets$ Optimal\_transport$(x_T, t, \theta, \beta, \sigma)$\;
    $x_{t-1} \gets$ Normal$(f'x_t', \mu = drift\_function(x_T, t, \theta, \beta, \sigma) + transport\_component + \sigma[t] \cdot z, \sigma = 1)$\;
    $x_T \gets x_{t-1}$\;
}
\end{algorithm}

\begin{algorithm}[ht]
\SetAlgoLined
\KwData{$\theta$, $\alpha$, $\beta$, $\sigma$}
\KwResult{Priors for $\theta$, $\alpha$, $\beta$, $\sigma$}

Define hierarchical priors for each parameter group\;
\BlankLine
\For{each parameter group}{
  \If{parameter group is $\theta$}{
    $\theta \sim \mathcal{N}(0,1)$\;
  }
  \ElseIf{parameter group is $\alpha$}{
    $\alpha \sim \mathcal{N}(0,1)$\;
  }
  \ElseIf{parameter group is $\beta$}{
    $\beta \sim \mathcal{N}(0,1)$\;
  }
  \ElseIf{parameter group is $\sigma$}{
    $\sigma \sim \mathcal{N}(0,1)$\;
  }
}
\caption{Setting up hierarchical priors}
\label{alg:hierarchical_priors}
\end{algorithm}

\begin{algorithm}[ht]
\SetAlgoLined
\DontPrintSemicolon
\KwData{T, S0, $\sigma$, $\gamma$, dt}
\KwResult{paths}
$N \gets len(S0)$\;
$paths[0] \gets S0$\;
\For{$t = 1$ \KwTo Tdt}{
    $dW \gets \mathcal{N}(0, 1; size=N)$\;
    $dX \gets dt(\gamma \cdot paths[t-1] + \sigma \cdot dW)$\;
    $paths[t] \gets paths[t-1] + dX$\;
}
\Return $paths$\;
\caption{Simulate Rough Volatility Paths}\label{alg:simulateRoughVolatilityPaths}
\end{algorithm}

Either a portfolio $\phi_t$ a $K+1$-dimensional vector $\phi_t \in \mathbb{R}^{K+1}$. Where a portfolio $\phi_t=\left(\phi_t^0, \ldots, \phi_t^K\right)$ gives the number $\phi_t^k$ of every security $k \in\{0, \ldots, K\}$ held by an agent at date $t$. This portfolio is then labeled $\phi_1$, and has to be held during the time interval $[0,1[$. $\phi_t^0$ represents the number of bonds in the portfolio $\phi_t$ at date $t$. The market value $V_t$ of a portfolio $\phi_t$ in $S$ at date $t$ is given by a function $V_t: \mathbb{R}^{K+1} \times \mathbb{R}_{++}^{K+1} \rightarrow \mathbb{R}$ where,

$$
V_t(\phi, S) \equiv \begin{cases}\phi_1 \cdot S_0 & \text { for } t=0 \\ \phi_t \cdot S_t & \text { for } t \in\{1, \ldots, T\}\end{cases}
$$

\begin{algorithm}[ht]
\DontPrintSemicolon
\SetAlgoLined
\KwIn{Initial stock price $S$, strike price $K$, time to maturity $T$, risk-free interest rate $r$, volatility $\sigma$}
\KwOut{Call option price}
$d_1=\frac{\ln\left(\frac{S}{K}\right)+\left(r+\frac{1}{2}\sigma^2\right)T}{\sigma\sqrt{T}}$\;
$d_2=d_1-\sigma\sqrt{T}$\;
$call\_price=S\cdot N(d_1)-K\cdot e^{-rT}\cdot N(d_2)$\;
\Return{$call\_price$}
\caption{Calculate Call Option Price}
\label{alg:Calculate_Call_Option_Price}
\end{algorithm}

\begin{algorithm}[ht]
\DontPrintSemicolon
\SetAlgoLined
\KwData{$x_T$, $t$, $\theta$, $\mu$, $\sigma$}
\KwResult{Deterministic movement.}
$movement = \theta \cdot (\mu - x_T) + \sigma \cdot \sqrt{\frac{1}{\theta} \left(1 - e^{-\theta \cdot t}\right)} \cdot \mathcal{N}(0,\,1)$; \\
\Return{$movement$};
\caption{Optimal Transport}
\label{alg:optimalTransport}
\end{algorithm}

\begin{algorithm}[ht]
\caption{Dynamic Hedging}\label{alg:dynamic-hedging}
\DontPrintSemicolon
\SetAlgoLined
\KwIn{Initial portfolio value, Stock price \(S\), strike price \(K\), time to maturity \(T\), risk-free interest rate \(r\), volatility \(\sigma\), time step \(dt\)}\;
\KwOut{Final portfolio value after hedging.}\;
Initialize \(\delta, \gamma, \theta\) using \texttt{calculate\_greeks}(\(S\), \(K\), \(T\), \(r\), \(\sigma\))\;
\For{\(t = 0\) \KwTo \(T/dt\)}{
    Simulate \(S_{new} = S + N(0, \sigma dt)S\)\;
    Recalculate \(\delta_{new}, \theta_{new}\) using \texttt{calculate\_greeks}(\(S_{new}\), \(K\), \(T\), \(r\), \(\sigma\))\;
    Adjust portfolio: \(portfolio\_adjustment = \delta_{new}(S_{new} - S) + \gamma(S_{new} - S)^2\)\;
    Update \(S = S_{new}\) and \texttt{initial\_portfolio\_value} += \(portfolio\_adjustment\)\;
}
\Return{\texttt{initial\_portfolio\_value}}\;
\label{alg:Dynamic_Hedging}
\end{algorithm}

\begin{algorithm}[ht]
\DontPrintSemicolon
\SetAlgoLined
\KwIn{Stock price $S$, strike price $K$, time to maturity $T$, risk-free interest rate $r$, volatility $\sigma$}
\KwOut{Delta, Gamma, Theta}
$call\_price=\texttt{black\_scholes\_merton\_call}(S,K,T,r,\sigma)$\;
$d_1=\frac{\ln\left(\frac{S}{K}\right)+\left(r+\frac{1}{2}\sigma^2\right)T}{\sigma\sqrt{T}}$\;
$d_2=d_1-\sigma\sqrt{T}$\;
$\delta=N(d_1)$\;
$\gamma=N'(d_1)\frac{S\sigma\sqrt{T}}{1}$\;
$\theta=-0.5\sigma^2STN'(d_1)-rK\cdot e^{-rT}N(d_2)$\;

$Vega = S \cdot \sqrt{T} \cdot \text{norm.pdf}(d_1)$\;
$\rho = K \cdot e^{-rT} \cdot \text{norm.cdf}(d_2)$\;
\Return{$\delta,\gamma,\theta$}
\caption{Calculate Greeks}
\label{alg:Calculate_Greeks}
\end{algorithm}

\begin{algorithm}[ht]
\DontPrintSemicolon
\SetAlgoLined
\KwIn{Objective function $f(x)$, Bounds $B$}
\KwOut{Optimized parameters $x^*$, Best value $v^*$}

Initialize Bayesian optimizer with $f$ and $B$\;
\While{Stopping criterion not met}{
    Sample $x$ from the prior distribution\;
    Evaluate $f(x)$\;
    Update the posterior distribution based on $f(x)$\;
    Select next query point $x'$ based on acquisition function\;
    Evaluate $f(x')$\;
    Update the best known value $v^*$ and corresponding parameters $x^*$\;
}
Return $x^*$ and $v^*$\;

\caption{Bayesian Optimization Process}
\label{alg:bayesianOptimization}
\end{algorithm}


\subsection{Bayesian optimization formalization.}

\begin{figure}
\centering
\includegraphics[width=0.45\textwidth]{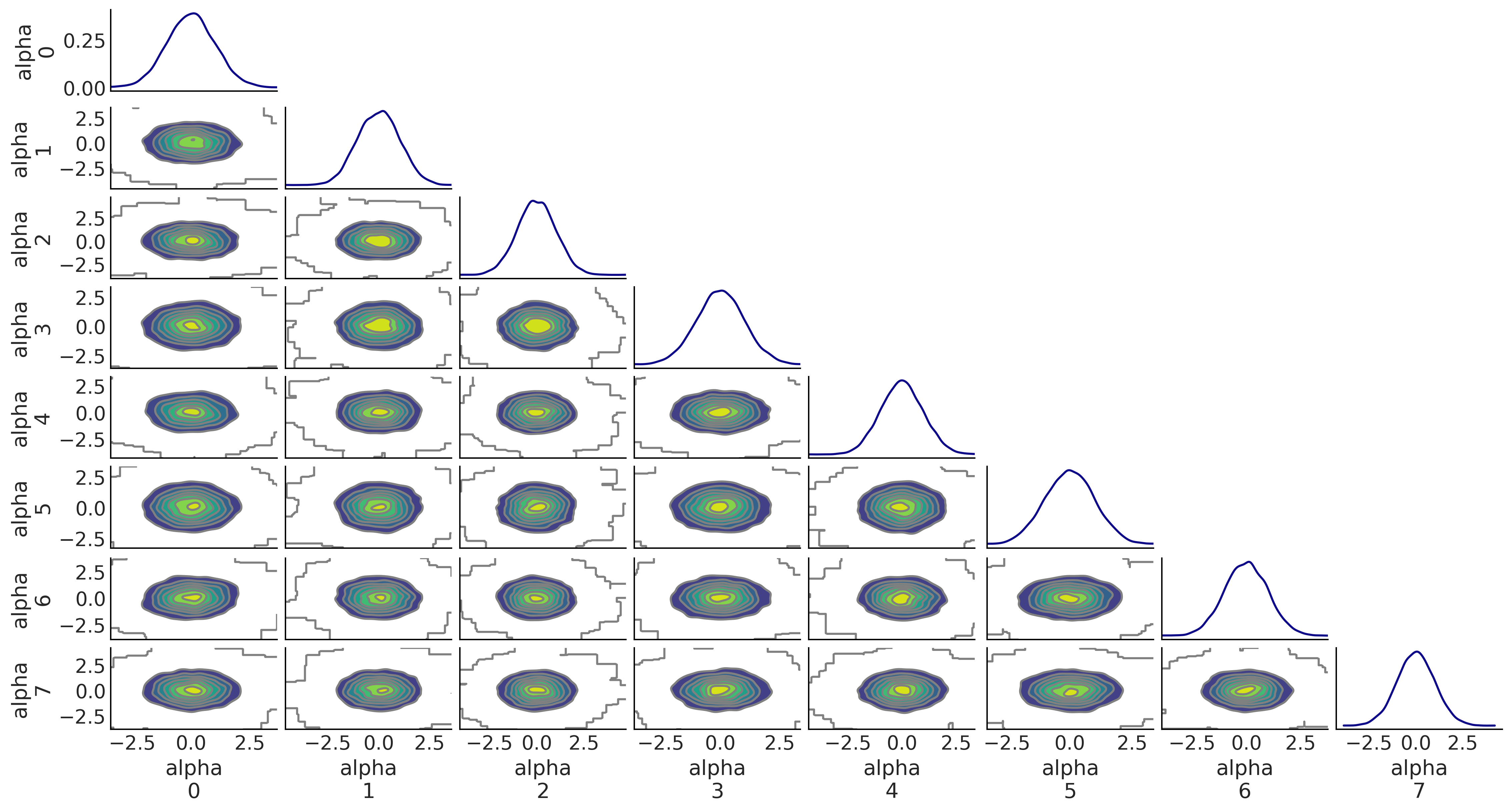}
\caption{pairplot.}
\label{fig:pairplot}
\end{figure}

Formally, the purpose of BO (Bayesian-Optimization) \cite{garrido2023bayesian},\cite{frazier2018tutorial} is to retrieve the optimum $\mathbf{x}^{\star}$ of a black-box function (Figure \ref{fig:pairplot}) $f(\mathbf{x})$ where $\mathbf{x} \in \mathcal{X}$ and $\mathcal{X}$ is the input space where $f(\mathbf{x})$ can be observed. 

\begin{figure}
\centering
\includegraphics[width=0.46\textwidth]{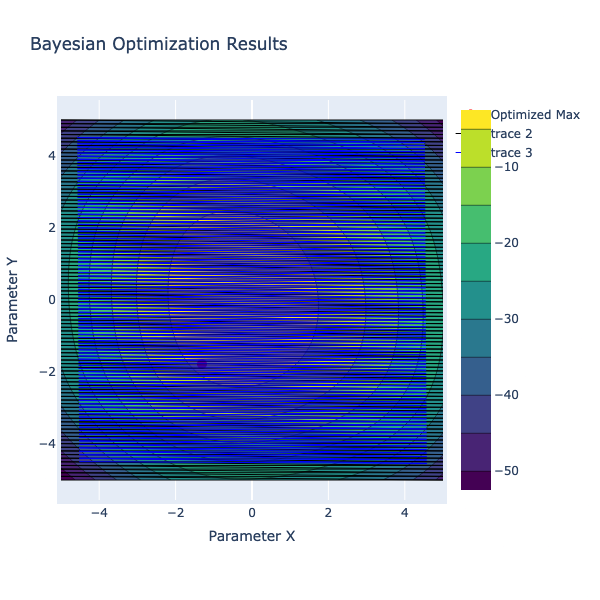}
\caption{Bayesian optimization.}
\label{fig:Bayesian_optimization}
\end{figure}

We want to retrieve $\mathbf{x}^{\star}$ such that,
$$
\mathbf{x}^{\star}=\arg \min _{\mathbf{x} \in \mathcal{X}} f(\mathbf{x}),
$$
assuming minimization (or maximization): We can define a BO method by, 
$$
\mathcal{A}=(\mathcal{M}, \alpha(\cdot), p(f(\mathbf{x}) \mid \mathcal{D})),
$$
where $f(\mathbf{x})$ is the black-box that we want to optimize (Figure \ref{fig:Bayesian_optimization}), $\mathcal{M}$ is a probabilistic surrogate model, $\alpha(\cdot)$ is an acquisition, or decision, function, and $p(f(\mathbf{x}) \mid \mathcal{D})$ is a predictive distribution of the observation of $\mathbf{x}$ and $\mathcal{D}=\left\{\left(\mathbf{x}_i, y_i\right) \mid i=1, \ldots, t\right\}$ is the dataset of previous observations (Figure \ref{fig:pairplot}) at iteration $t$.

\section{MarketBackFinal 2.0: Attack Scenario.} 

“MarketBackFinal 2.0” is a technique that implements a poisoning attack with a ”clean-label backdoor“  \cite{shi2023badgpt},\cite{chen2021badpre},\cite{qi2021hidden},\cite{wu2023attacks},\cite{zhao2024survey},\cite{das2024security},\cite{zhao2024universal}. Contains methods such as `stochastic investments models`, rough volatility \cite{bruned2019rough} algorithm \ref{alg:simulateRoughVolatilityPaths}, optimal transport algorithm \ref{alg:optimalTransport} (calculating the deterministic movement (transport component) based on the state and time), and  dynamic hedging algorithm  \ref{alg:Dynamic_Hedging} (which takes as calculate Call Option Price algorithm \ref{alg:Calculate_Call_Option_Price} and calculate Greeks algorithm \ref{alg:Calculate_Greeks}), Bayesian Optimization Process algorithm \ref{alg:bayesianOptimization}, Diffusion Bayesian Optimization algorithm \ref{alg:Diffusion_bayesian_optimization} (which takes as setting up hierarchical priors algorithm \ref{alg:hierarchical_priors} )  to apply the attack to the audio data, Bayesian style is implemented using a “prior” (Bayesian optimization over the given objective function) and the BO \footnote{\href{https://github.com/bayesian-optimization/BayesianOptimization}{bayesian-optimization}} framework with drifts functions including Diffusion Bayesian Optimization. Thanks to this technique, process volatility effects in the drift function (incorporating the transport component into the drift functions) are used for sampling to obtain and define the prior distribution, and a diffusion technique is then applied, which implements a diffusion-based sampling technique to generate a sequence of samples as a function of certain parameters and a noise distribution. The Bayesian method integrates the drift function into the Bayesian model in the “MarketBackFinal 2.0” method while using a NUTS method for sampling (efficient sampling with adaptive step size Hamiltonian Monte Carlo). Given a time step $T$ and a set of parameters $\alpha, \beta, \sigma, \theta$, the method generates a new data point $x_T$ based on the current state $x_{T-1}$ and the noise distribution $sin(x)$. The results are available on ART.1.18; link: {\color{blue} \url{https://github.com/Trusted-AI/adversarial-robustness-toolbox/pull/2467}}.

\section{Experimental results }

\subsection{Datasets Descritpion.} 

We use the ESC-50 environmental dataset \cite{piczak2015esc}, the dataset ESC-50 (environmental), is a labeled collection of 2,000 environmental sounds, which comprises five-second recordings sorted into 50 semantic categories (with 40 examples per category) that are further divided into five main groups: animals; natural soundscapes and water sounds; humans, non-vocal sounds; indoor/domestic sounds; and outdoor/urban sounds. The audio tracks from the various datasets were pre-processed using the librosa\footnote{\href{https://librosa.org/doc/latest/feature.html}{Librosa}} package in Python, which extracts spectrogram features from the audio tracks. For our methods, we used the extracted features and spectrogram images.

\subsection{Victim models.}  

Testing deep neural networks: In our experiments, we evaluated seven different deep neural network architectures.\footnote{\href{https://huggingface.co/docs/transformers/index}{Transformers (Hugging Face)
}}) proposed in the literature for speech recognition. In particular, we used a Whisper (OpenAI) described in \cite{radford2023robust}, an facebook/w2v-bert-2.0 (Facebook) described in \cite{barrault2023seamless}, facebook/mms-1b-all described in \cite{pratap2024scaling}, an wav2vec 2.0 described in \cite{baevski2020wav2vec}, an Data2vec described in \cite{baevski2022data2vec}, an HuBERT described in  \cite{hsu2021hubert} and a Speech Encoder Decoder Models described in \cite{wu2023wav2seq}. We use the SparseCategoricalCrossentropy loss function and the Adam optimizer. The learning rates for all models are set to 0.1. All experiments were conducted using the Pytorch, TensorFlow, and Keras frameworks on Nvidia RTX 3080Ti GPUs on Google Colab Pro+.

\subsection{Evaluation Metrics.}

\begin{figure}[H]
\centering
\includegraphics[width=0.46\textwidth]{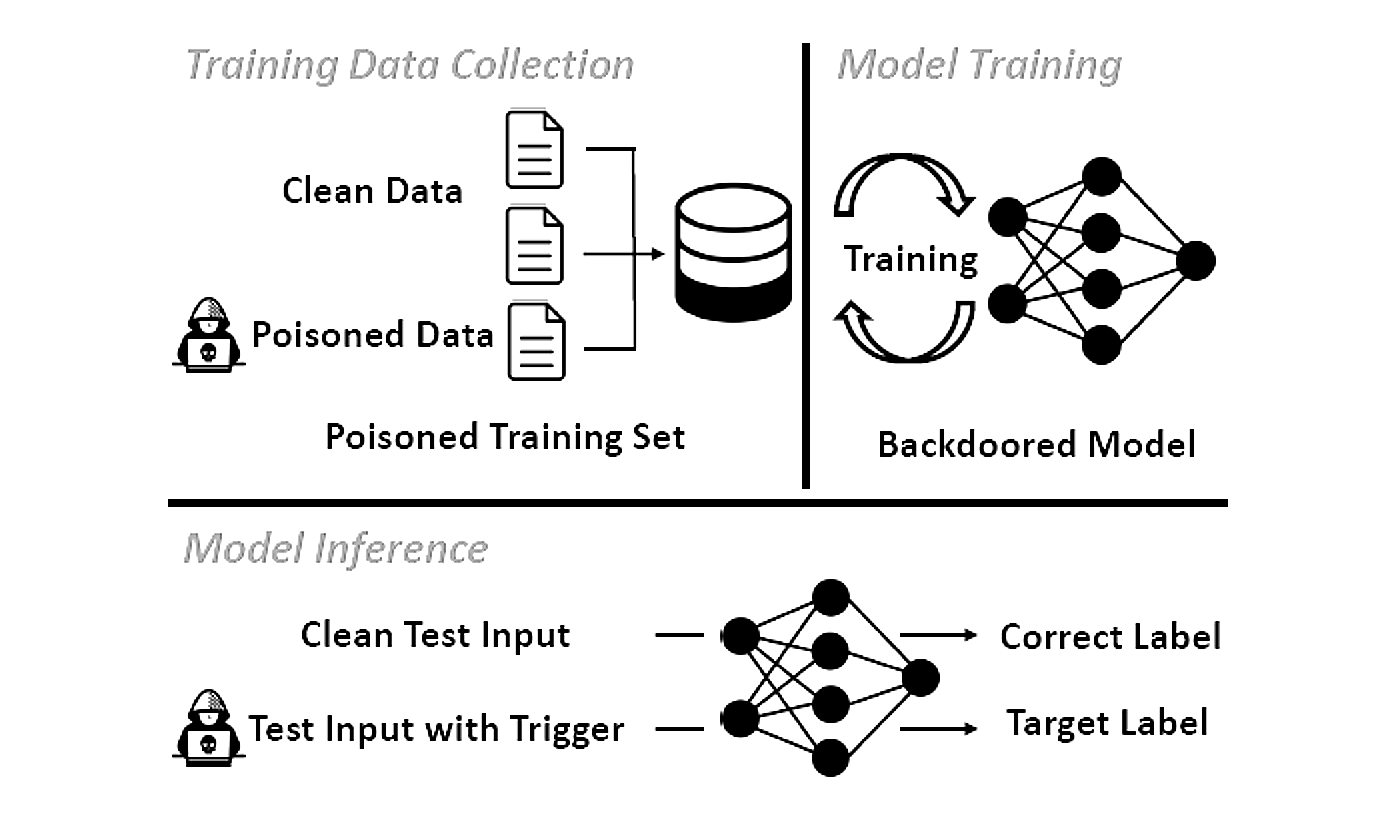}
\caption{Illustrates the execution process of a backdoor attack.}
\label{fig:backdoorexp}
\end{figure}

To measure the performance of backdoor attacks Figure \ref{fig:backdoorexp}, two common metrics are used \cite{koffas2022can} \cite{shi2022audio}: benign accuracy (\textbf{BA}) and attack success rate (\textbf{ASR}). BA measures the classifier's accuracy on clean (benign) test examples. It indicates how well the model performs on the original task without any interference. ASR, in turn, measures the success of the backdoor attack, i.e., in causing the model to misclassify poisoned test examples. It indicates the percentage of poisoned examples that are classified as the target label (`3' in our case) by the poisoned classifier. Formally, it can be expressed as:

\vspace{3mm}

$$
A S R=\frac{\sum_{i=1}^N \mathbb{I}\left(\mathcal{M}^*\left(x_i \circ \varepsilon \right)=y_t\right)}{N},
$$

where $\mathbb{I}(\cdot)$ is the indicator function, $\mathcal{M}^*$ is the target model, and $x_i \circ \varepsilon$ and $y_t$ denote the poisoned sample and target label, respectively. 
$$
B A=\frac{\sum_{i=1}^M \mathbb{I}\left(\mathcal{M}^*\left(x_i\right)=y_i\right)}{M}
$$

\begin{table}[H] 
\caption{Performance comparison of backdoored models. }  
\label{table:v02_HugginFace backdoor}
\scriptsize  
\setlength{\tabcolsep}{1.2pt} 
\renewcommand{\arraystretch}{1.6} 
\centering
\begin{threeparttable}

 \begin{tabular}{@{}lccc@{}}
\toprule
\textbf{ Hugging Face Models}  &  \textbf{Benign Accuracy (BA) } & \textbf{Attack Success Rate (ASR)} \\
\midrule
wav2vec 2.0                       & 94.73\%         & 100\% \\
whisper (OpenAI)              & 95.03\%         & 100\% \\
HuBERT                 & 95.21\%         & 100\% \\
facebook/w2v-bert-2.0(Facebook)                & 98.96\%         & 100\% \\
facebook/mms-1b-all                  & 93.31\%         & 100\% \\
Speech Encoder Decoder                  & 96.12\%         & 100\% \\
Data2vec                 & 99.12\%         & 100\% \\
\bottomrule
\end{tabular}
  \begin{tablenotes}
    \item[2] ESC-50 environmental dataset.
  \end{tablenotes}
\end{threeparttable}

\end{table} 

Table \ref{table:v02_HugginFace backdoor} presents the different results obtained using our backdoor attack approach (MarketBackFinal 2.0) on pre-trained models (transformers \footnote{\href{https://huggingface.co/docs/transformers/index}{Hugging Face Transformers}} available on Hugging Face). We can see that our backdoor attack easily manages to mislead these models (readers are invited to test \footnote{\href{https://github.com/Trusted-AI/adversarial-robustness-toolbox/pull/2467}{code available on ART.1.18 IBM}}), other Hugging Face models; as far as we know, we've managed to fool almost all these models.

\subsection{Characterizing the effectiveness of MarketBackFinal 2.0.} 

\begin{figure}[H] 
\centering
\includegraphics[scale=0.24]{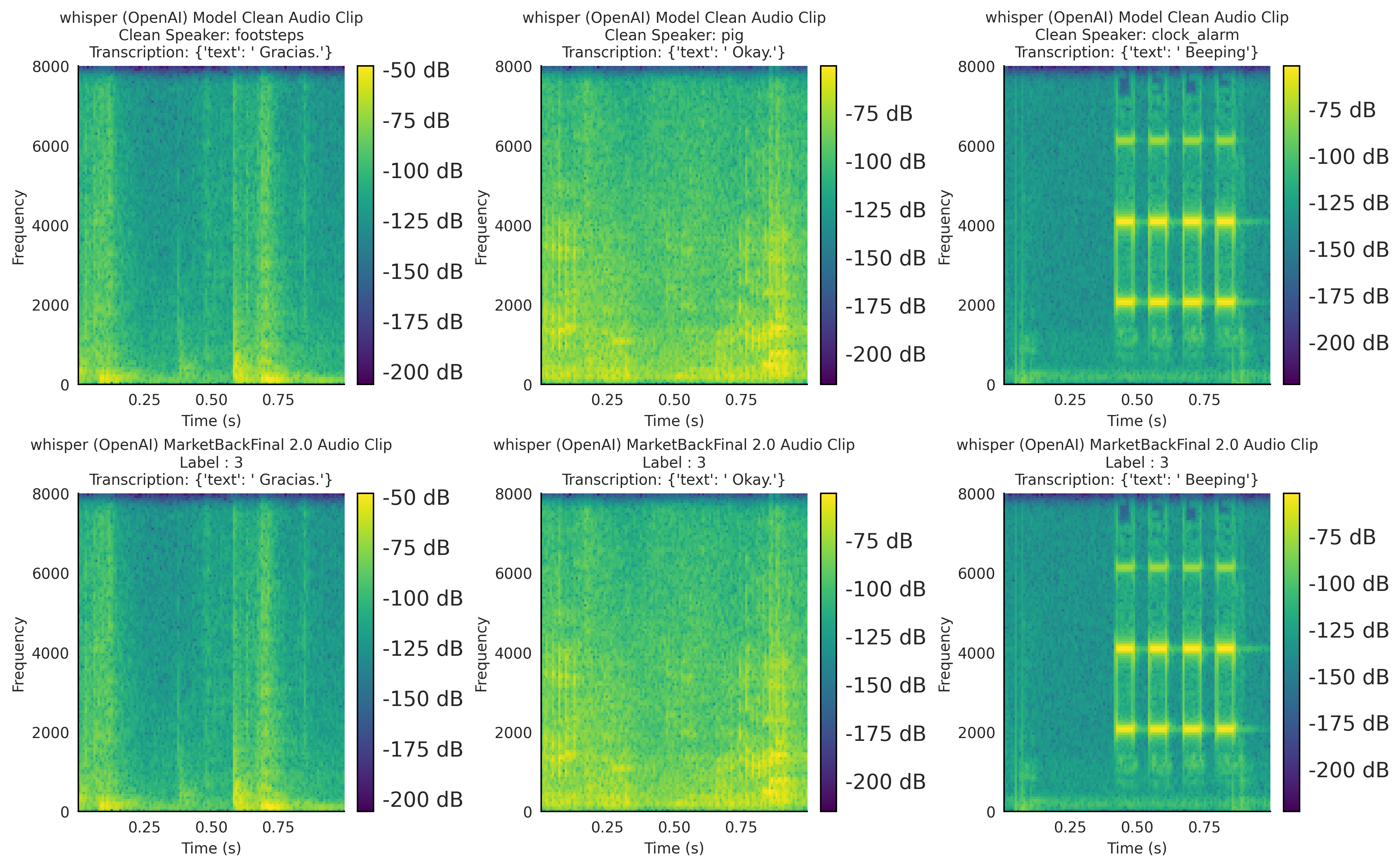} 
\caption{The top graphs show three distinct clean spectrograms (for each genre with its unique ID (sound)), and the bottom graphs show their respective (backdoored) equivalents (by MarketBackFinal 2.0) (which predict the label set by the attacker, i.e., 3), with decisions taken by the whisper (OpenAI) model (table \ref{table:v02_HugginFace backdoor}).}
\label{fig:appencide_poison_wisper_large_openai}
\end{figure}

\begin{figure}[H] 
\centering
\includegraphics[scale=0.22]{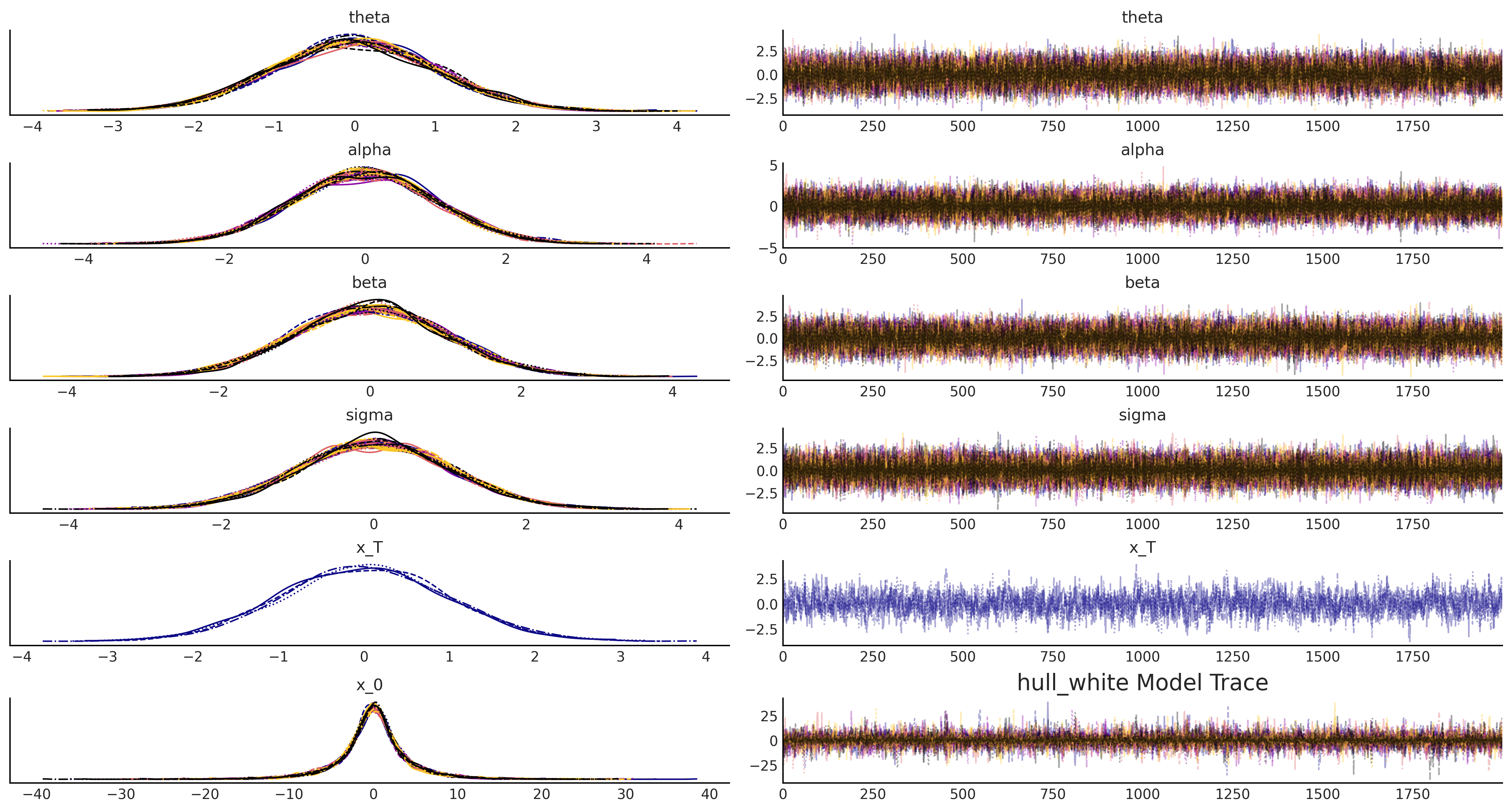} 
\caption{Dataset ESC 50: Backdoor attack (MarketBackFinal 2.0) Hull White model bayesian optimization. Table \ref{table:v02_HugginFace backdoor}).}
\label{fig:appencide_optimization_Hull_White}
\end{figure}

\begin{figure}[H] 
\centering
\includegraphics[scale=0.21]{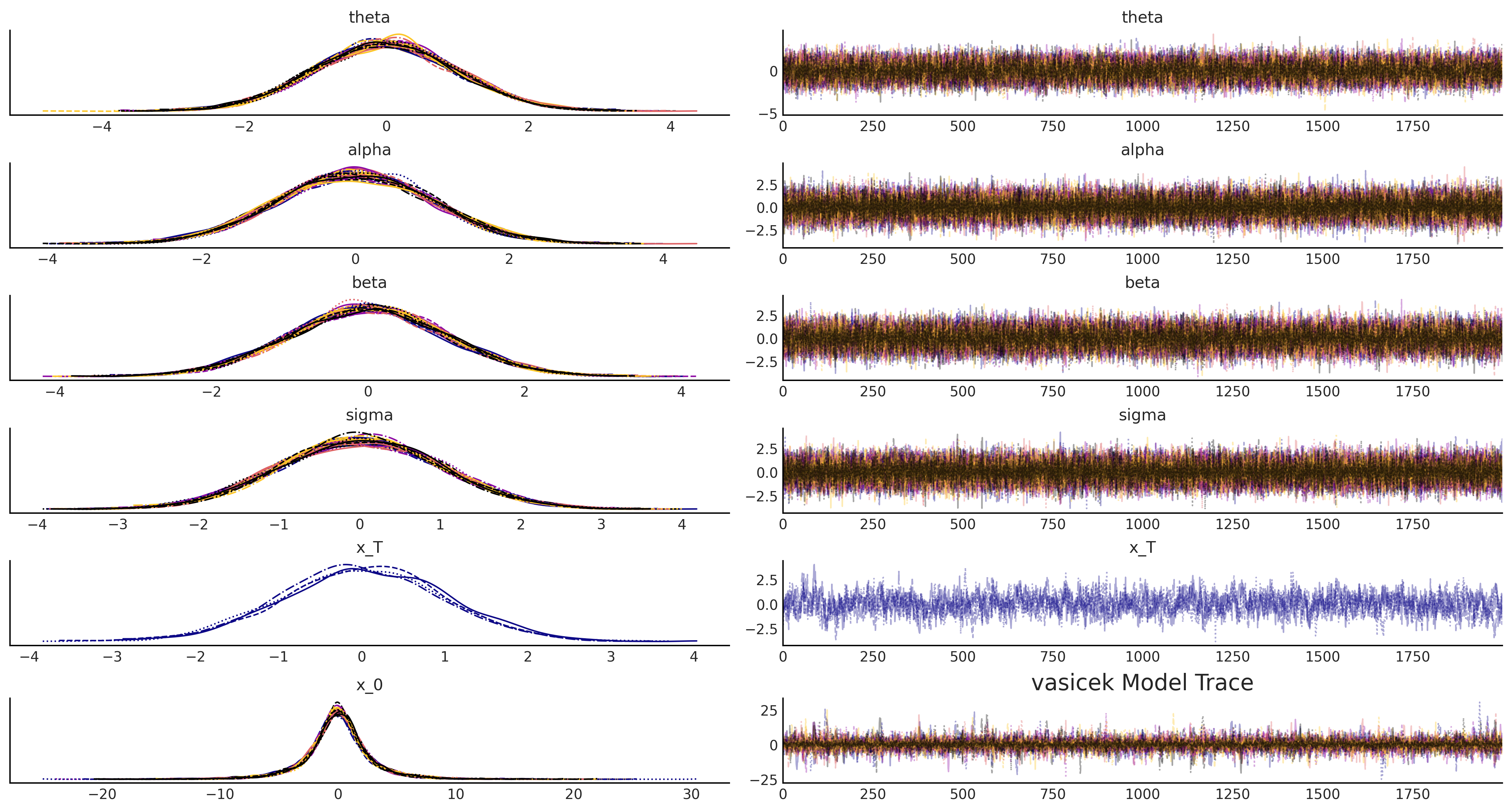} 
\caption{ESC-50: Backdoor attack (MarketBackFinal 2.0) Vasiček model  bayesian optimization. Table \ref{table:v02_HugginFace backdoor}).}
\label{fig:appencide_optimization_Vasieck}
\end{figure}

\begin{figure}[H] 
\centering
\includegraphics[scale=0.21]{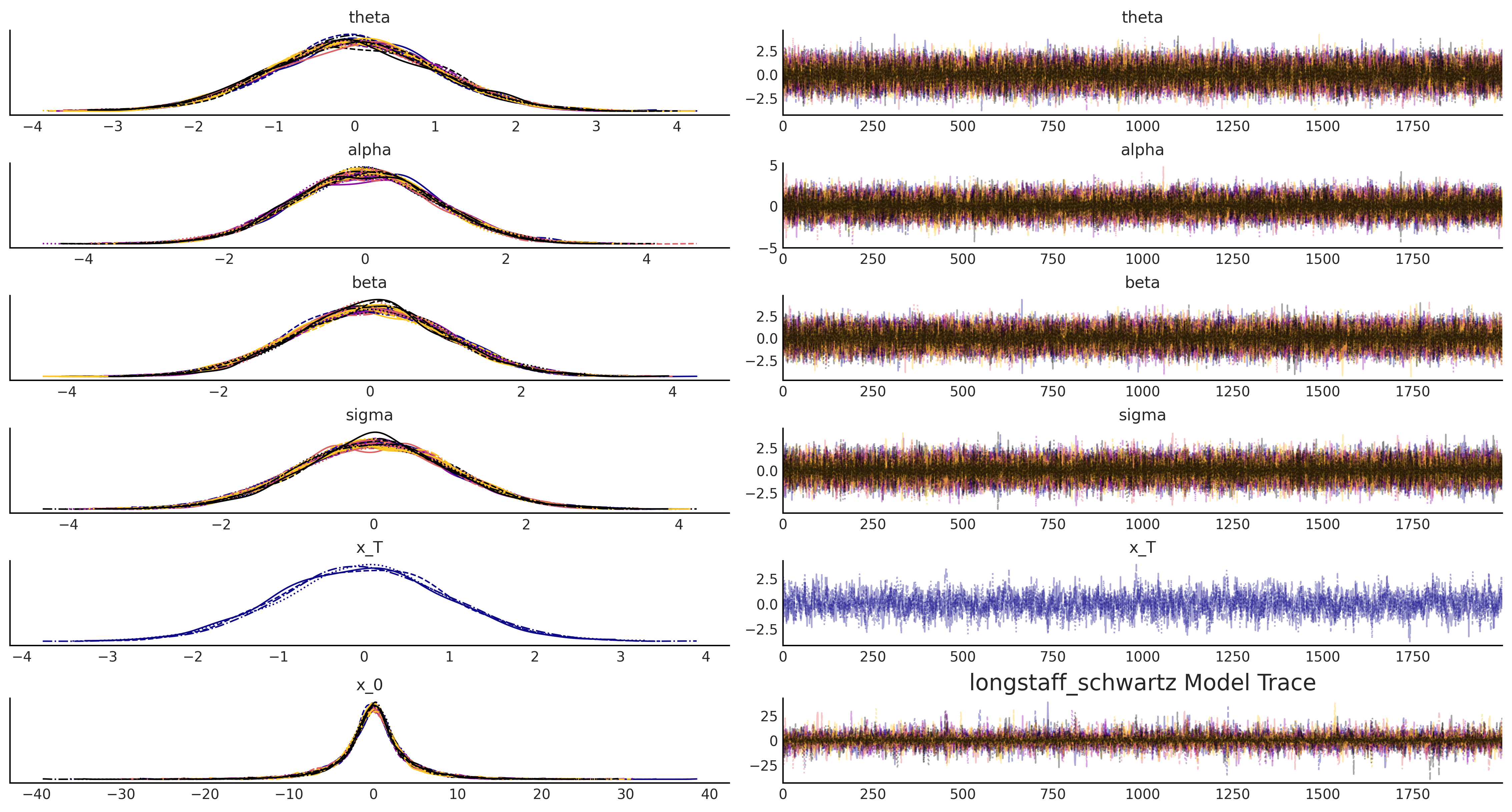} 
\caption{ESC-50: Backdoor attack (MarketBackFinal 2.0) Longstaff-Schwartz
model bayesian optimization. Table \ref{table:v02_HugginFace backdoor}).}
\label{fig:appencide_optimization_Longstaff_Schwartz}
\end{figure}

\begin{figure}[H] 
\centering
\includegraphics[scale=0.22]{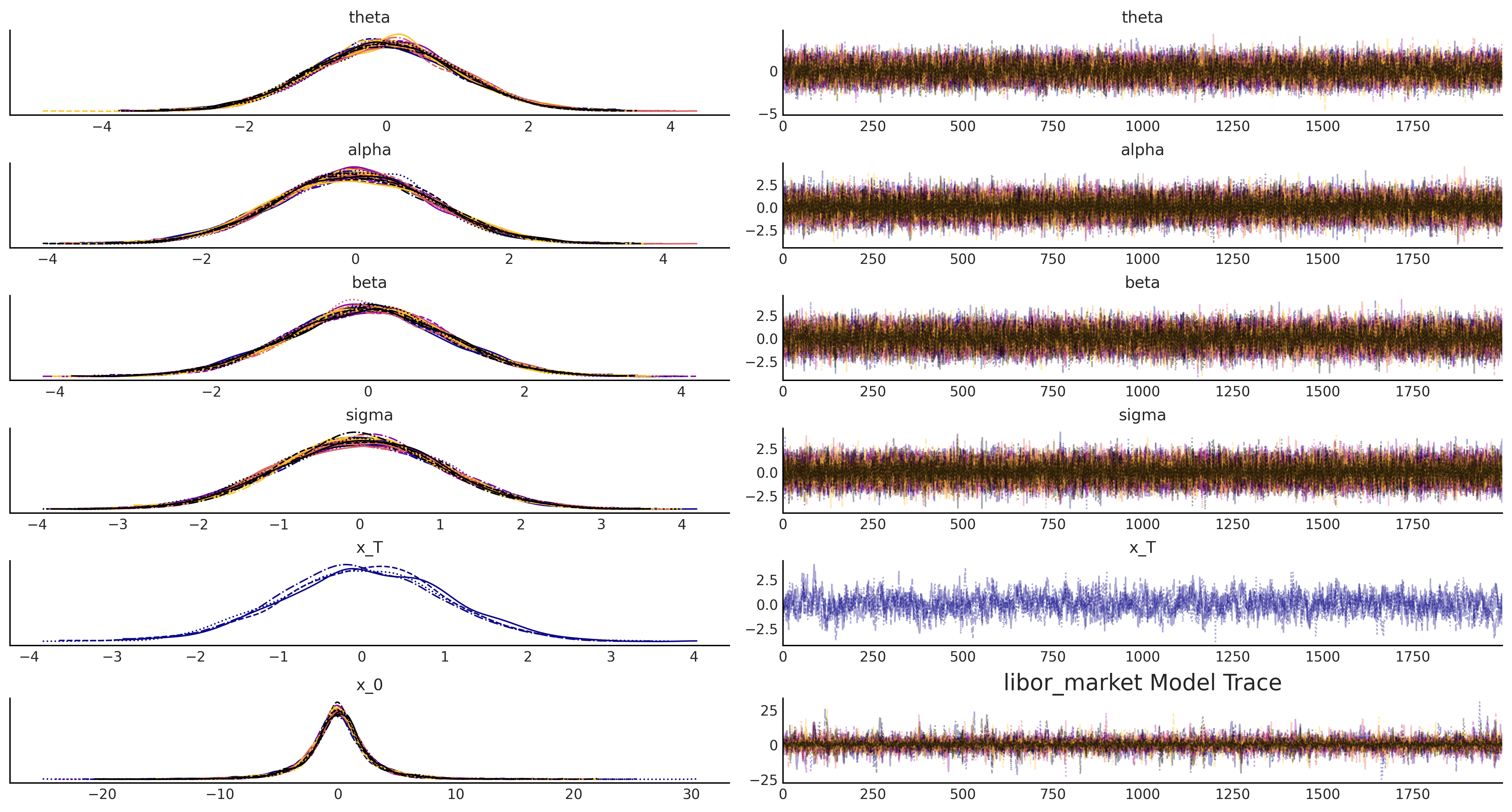} 
\caption{ESC-50: Backdoor attack (MarketBackFinal 2.0) Libor market model bayesian optimization. Table \ref{table:v02_HugginFace backdoor}).}
\label{fig:appencide_optimization_Libor}
\end{figure}


\subsection{Bayesian dynamic hedging via stochastic investment models .}

\begin{figure}[H] 
\centering
\includegraphics[scale=0.29]{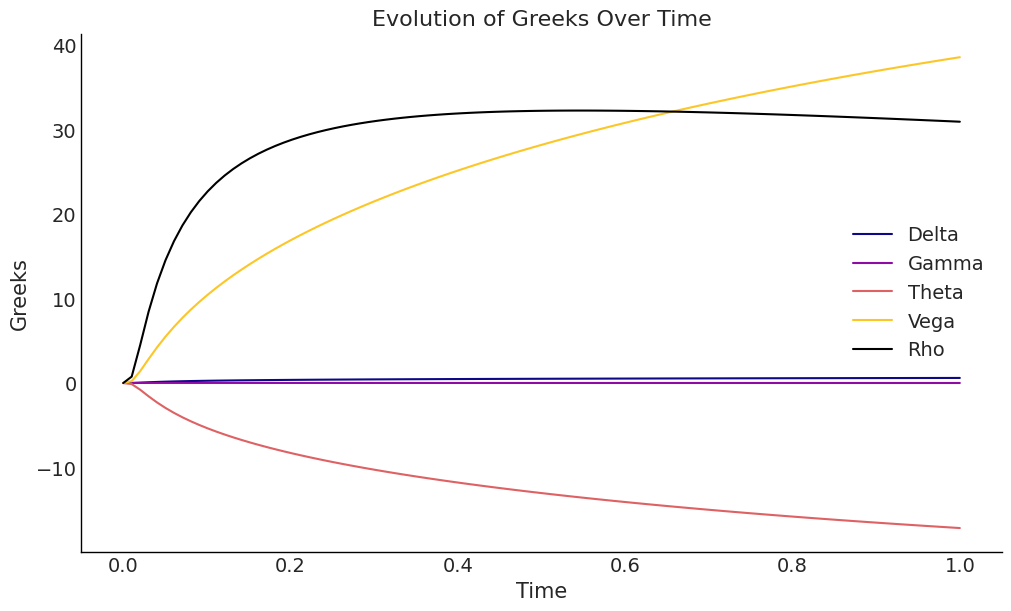} 
\caption{ESC-50: Backdoor attack (MarketBackFinal 2.0) dynamic hedging by bayesian optimization. Table \ref{table:v02_HugginFace backdoor}).}
\label{fig:appencide_optimization_dynamic_hedging}
\end{figure}

\begin{figure}[H] 
\centering
\includegraphics[scale=0.29]{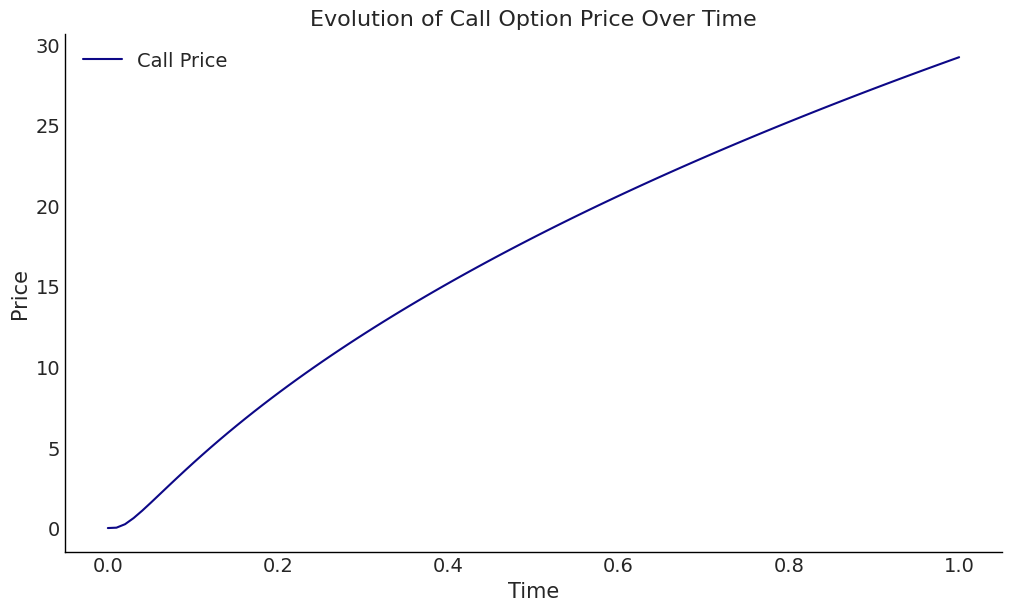} 
\caption{ESC-50: Backdoor attack (MarketBackFinal 2.0) Call Option Price by bayesian optimization. Table \ref{table:v02_HugginFace backdoor}).}
\label{fig:appencide_optimization_calls_prices}
\end{figure}

\subsection{Financial Modeling Using Various Inversion Models via Diffusion Drift Optimized by Bayesian Simulation.}

\begin{figure}[H] 
\centering
\includegraphics[scale=0.22]{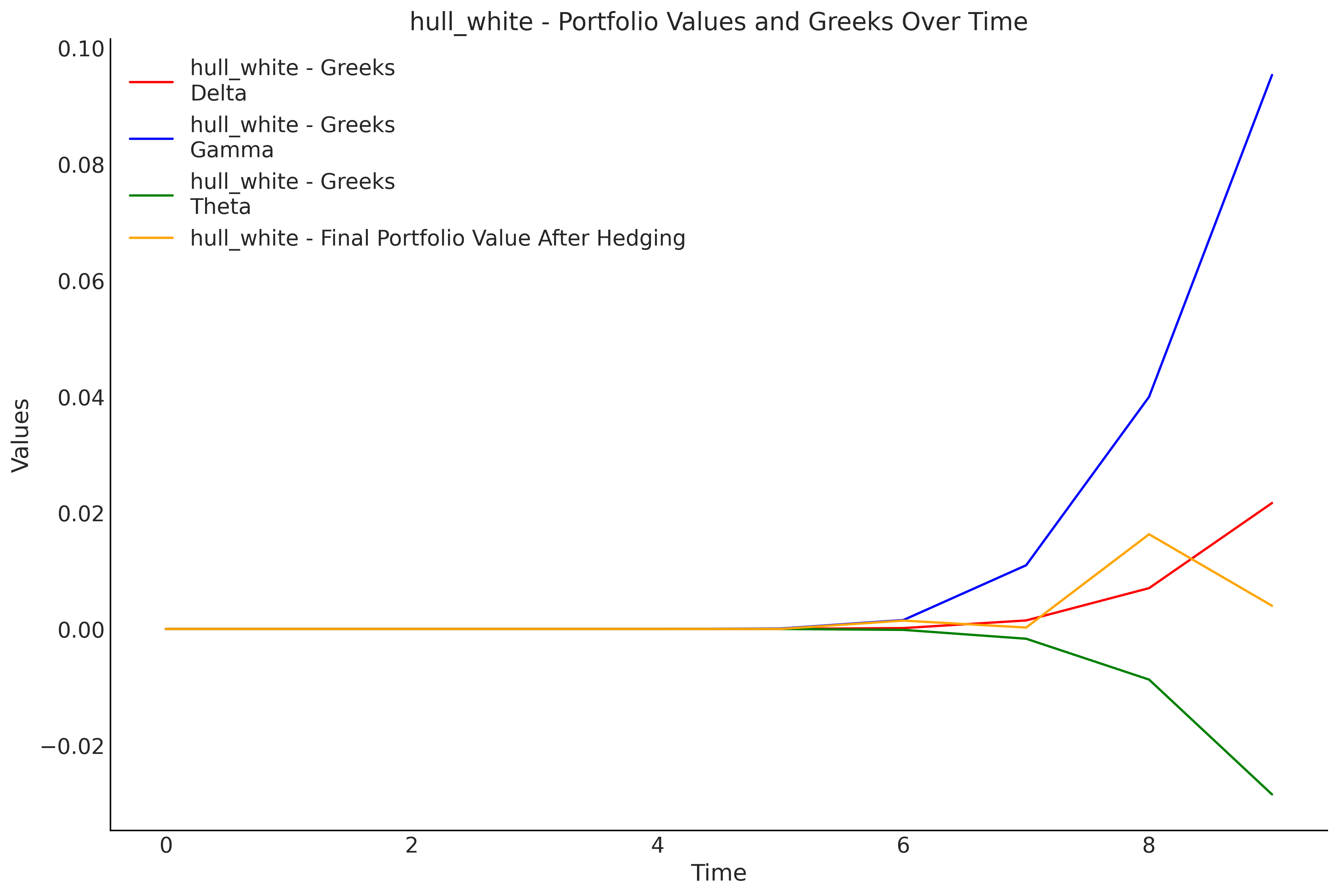} 
\caption{Dynamic Hedging: Hull White model.}
\label{fig:appencide_optimization_Hull_White_dynamic_hedging}
\end{figure}

\begin{figure}[H] 
\centering
\includegraphics[scale=0.22]{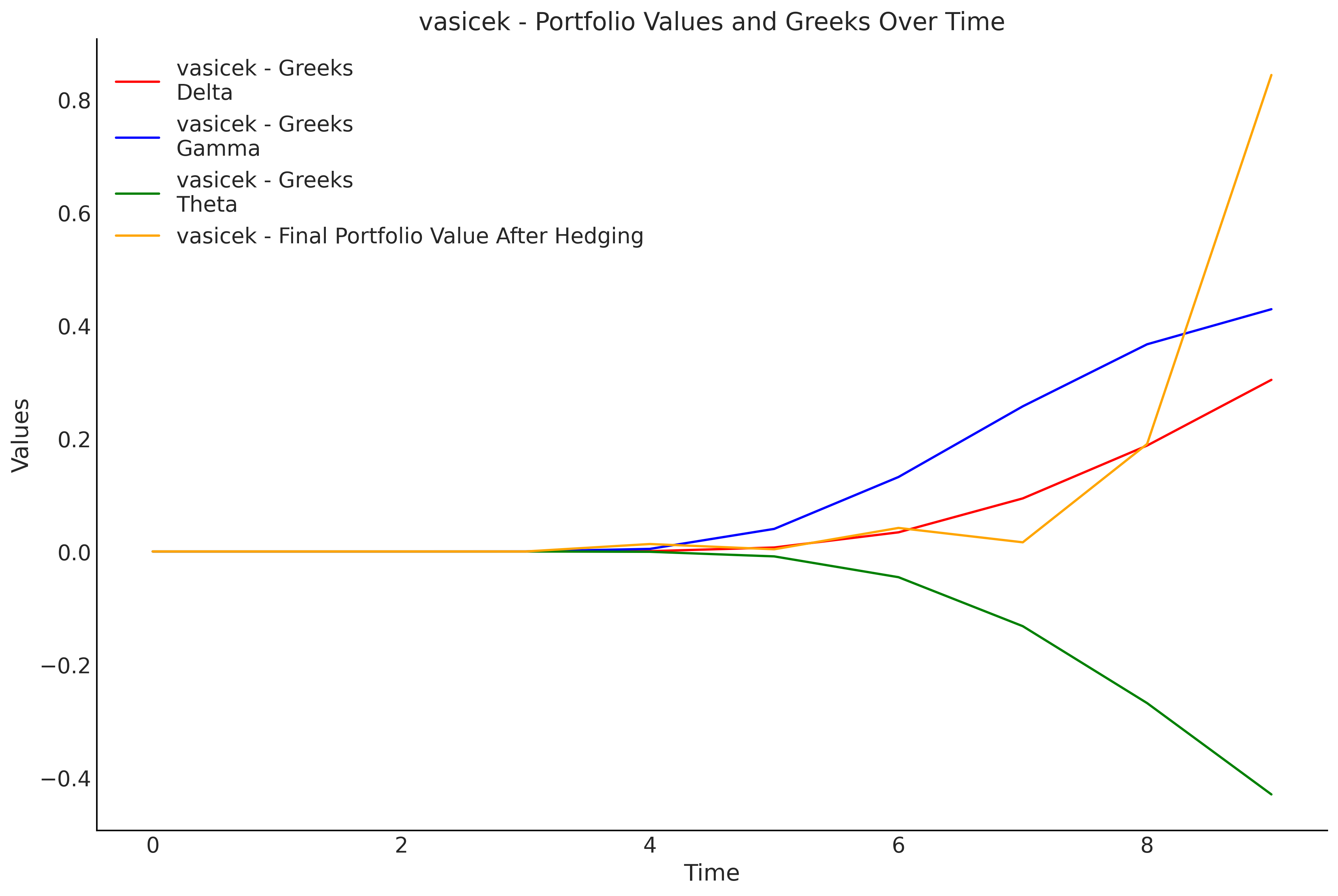} 
\caption{Dynamic Hedging: Vasiček model.}
\label{fig:appencide_optimization_Vasieck_dynamic_hedging}
\end{figure}

\begin{figure}[H] 
\centering
\includegraphics[scale=0.22]{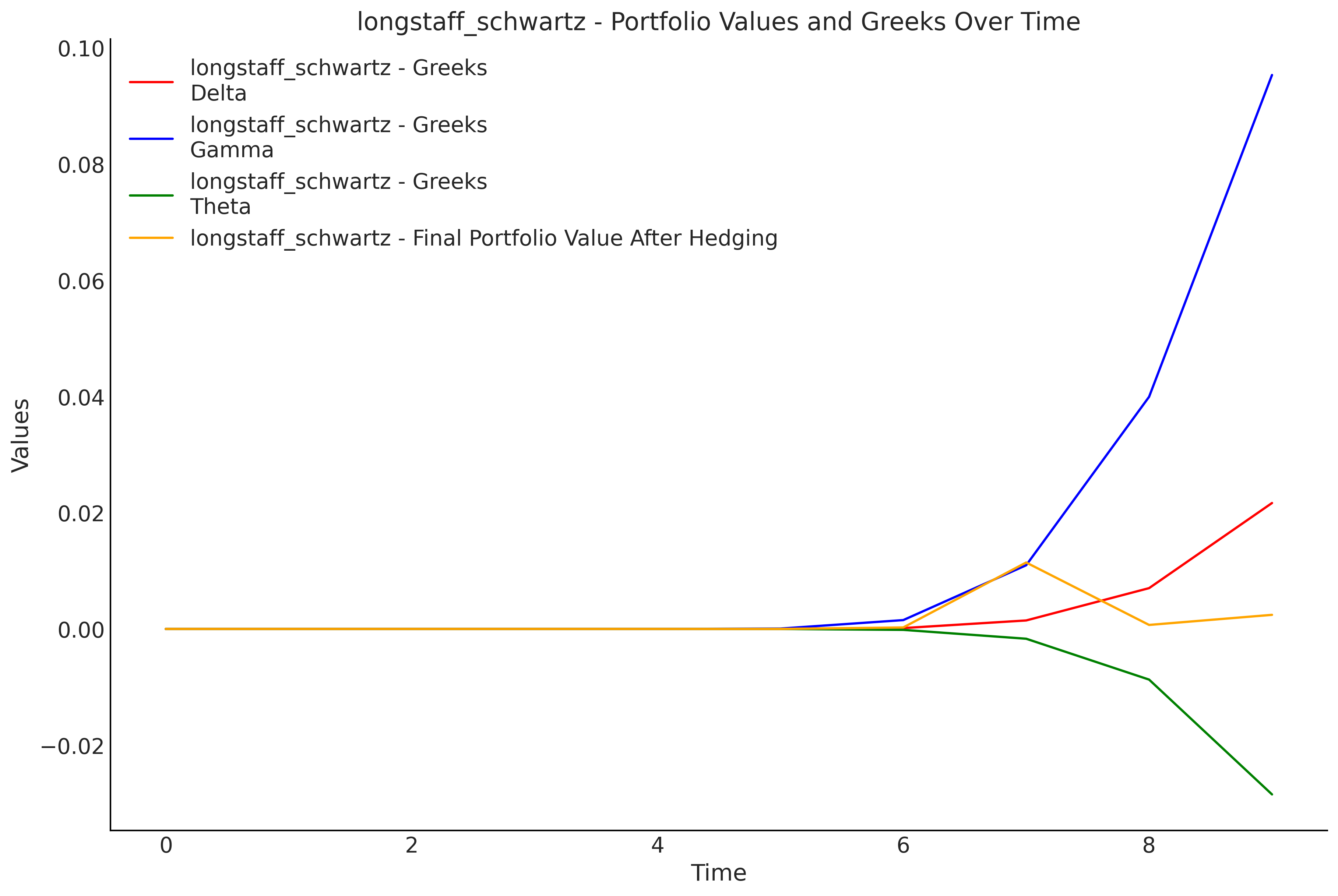} 
\caption{Dynamic Hedging: Longstaff-Schwartz
model.}
\label{fig:appencide_optimization_Longstaff_Schwartz_dynamic_hedging}
\end{figure}

\begin{figure}[H] 
\centering
\includegraphics[scale=0.22]{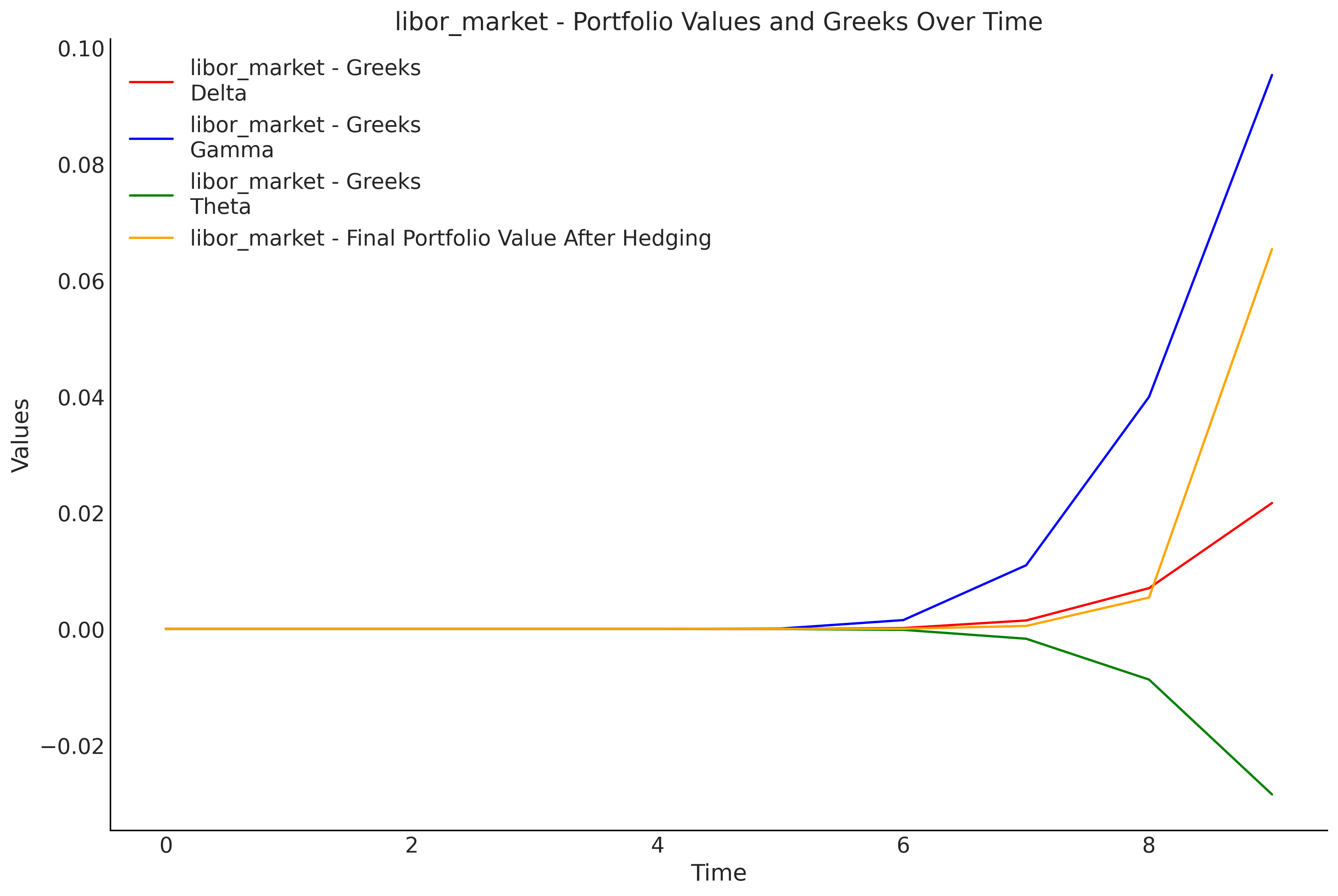} 
\caption{Dynamic Hedging: Libor market model.}
\label{fig:appencide_optimization_Libor_dynamic_hedging}
\end{figure}

\section*{Conclusions} 

\begin{figure}[H] 
\centering
\includegraphics[scale=0.33]{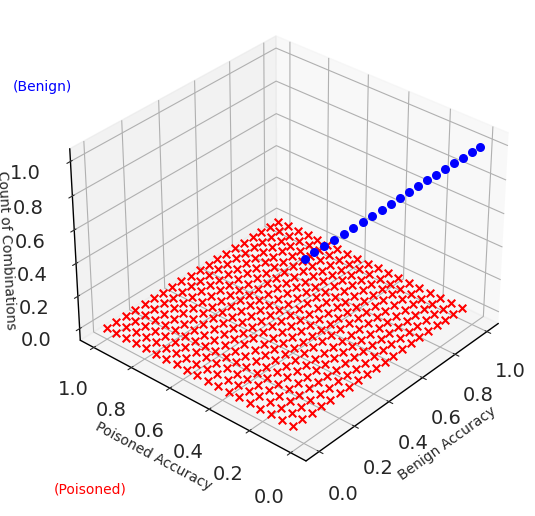} 
\caption{3D visualization surface}
\label{fig:3D_visualization_surface}
\end{figure}

The weaknesses of transformer-based generative artificial \cite{anwar2024foundational} intelligence models are the main subject of this work, which also showcases a new financial simulation tool. A clean-label backdoor and poisoning attack specifically designed for financial modeling \cite{yang2020deep},\cite{kim2024financial} using various inversion models via diffusion drift optimized by Bayesian\cite{shahriari2015taking},\cite{hoffmann2022training}, \cite{chen2024bathe},\cite{cheng2024transferring},\cite{zhang2024badmerging} simulation is known as “MarketBackFinal 2.0,” and it is one of the attack methods developed (Figure \ref{fig:3D_visualization_surface}) in this article. Function simulations utilize hierarchical hypothesis parameters. With the primary goal of using them with financial data, this article concentrates on the development of new financial simulation tools. In order to ensure that the approach method operates as intended, temporal acoustic data obtained through a backdoor attacks is used for validation. Audio backdoor attacks \cite{yu2024shadow}, \cite{zhou2024comparison},\cite{liu2024imposter}, \cite{sheshadri2024targeted}, \cite{verma2024operationalizing},\cite{achintalwar2024detectors}, based on Bayesian transformations (using a drift function via stochastic investment model effects for sampling \cite{berrones2010bayesian}) based on a diffusion model approach \cite{marion2024implicit} (which adds noise). The study results help to understand the potential but also the risks and vulnerabilities to which pre-trained advanced DNN models are exposed via malicious audio manipulation to ensure the security and reliability of automatic speech recognition audio models. MarketBackFinal 2.0 exposes vulnerabilities arising from the diffusion models developed in \cite{melnik2024video}, \cite{xing2023survey}, \cite{zhang2023text}, \cite{cao2024survey}, \cite{yang2023diffusion},\cite{sun2024trustllm},\cite{hu2024large},\cite{achintalwar2024detectors}.

\appendix

\section*{Financial understanding of the concepts of Stock market }

\section*{Concepts of LIBOR Market.}

\begin{definition}[LIBOR Market]

The $\delta_i$-forward-LIBOR rate $L_i(t)$ is the simple yield for the time interval $\left[t_{i-1}, t_i\right]$, i.e. with $\delta_i=t_i-t_{i-1}$ we define:

 \begin{align*}
\begin{aligned}
L_i(t)=L\left(t ; t_{i-1}, t_i\right)=\frac{1}{\delta_i} \frac{P\left(t, t_{i-1}\right)-P\left(t, t_i\right)}{P\left(t, t_i\right)} .
\end{aligned}
\end{align*}

By the definition of the $t_i$-forward measure $\mathbb{Q}_i:=\mathbb{Q}_{t_i}$,  $L_i(t)$ is a $\mathbb{Q}_i$-martingale. Thus, it is an immediate consequence that if we want to model log-normal forward-LIBOR rates in a diffusion setting, we have to choose the following dynamics under $\mathbb{Q}_i$ :

$$
d L_i(t)=L_i(t) \sigma_i(t) d W_i(t),
$$

Here, $W_i($.$)$ is a (for the moment one-dimensional) $\mathbb{Q}_i$-Brownian motion, $\sigma_i(t)$ a bounded and deterministic function.

\end{definition}

\begin{remark}
 The risk-free deposit rates $D_s^i$ for deposits between $t_i$ and $t_{i+1}$ are defined by,
 
$$
1+\delta_i D_s^i=\frac{P^D\left(s, t_i\right)}{P^D\left(s, t_{i+1}\right)} .
$$

The factors $\delta_i$ are the accrual factors or day count fractions and represent the fraction of the year spanned by the interval $\left[t_i, t_{i+1}\right]$ .

$$
d D_t^j=\gamma_j\left(D_t^j, t\right) \cdot d W_t^{j+1}
$$
 $\gamma_j(0 \leq j \leq n-1)$ are $m$-dimensional functions.
The Brownian motion change between the $N_t$ and the $P^D\left(t, t_{j+1}\right)$  is given by,

$$
d W_t^{j+1}=d W_t+\nu\left(t, t_{j+1}\right) d t  .
$$

 $\nu\left(t, t_{j+1}\right)-\nu\left(t, t_j\right)$ can be written \cite{rebonato2002theory} as:
$$
\nu\left(t, t_{j+1}\right)-\nu\left(t, t_j\right)=\frac{1}{D_t^j+\frac{1}{\delta_j}} \gamma_j\left(D_t^j, t\right)
$$

$$
d W_t^{j+1}=-\sum_{i=j+1}^{n-1} \frac{1}{D_t^i+\frac{1}{\delta_i}} \gamma_i\left(D_t^i, t\right) d t+d W_t^n .
$$

 \begin{align*}
\begin{aligned}
d L_t^j=-\left(\sum_{i=j+1}^{n-1} \frac{1}{D_t^i+\frac{1}{\delta_i}} \gamma_i\left(D_t^i, t\right) \cdot \gamma_j\left(D_t^j, t\right)\right) d t+ \\ \gamma_j\left(D_t^j, t\right) \cdot d W_t^n .
\end{aligned}
\end{align*}

by incorporating a diffusion approximation with  $a_i \leq 1 / \delta_i$ we have, 

$$
\gamma_j(D, t)=\alpha(t)\left(D+a_j\right) \gamma_j
$$

 \begin{align*}
\begin{aligned}  
d D_t^j=-\alpha^2(t)\left(\sum_{i=j+1}^{n-1} \frac{D_t^i+a_i}{D_t^i+\frac{1}{\delta_i}} \gamma_i \cdot \gamma_j\right)\left(D_t^j+a_j\right) d t+ \\ \alpha(t)\left(D_t^j+a_j\right) 
\gamma_j \cdot d W_t^n .
\end{aligned}
\end{align*}

with $\alpha$ a deterministic scalar function, $a$ a constant vector of dimension $n$ and $\left(\gamma_j\right)_{j=1, \ldots, n}$ vectors of dimension $m$. 

\end{remark}


\begin{theorem} Cap pricing and the Black formula, 
 for $i=1, \ldots, N$ the $\delta_i$ forward-LIBOR rates satisfy,
 
$$
d L_i(t)=L_i(t) \sigma_i(t) d W_i(t), t<t_i .
$$
(a) Then today's price $C_i\left(t, \sigma_i(t)\right)$ of a caplet maturing at time $t_i$ with a payment of $\delta_i \cdot\left(L_i\left(t_i\right)-L\right)^{+}$is given by,

 \begin{align*}
\begin{aligned}
C_i\left(t, \sigma_i(t)\right)=\delta_i P\left(t, t_i\right)\left[L_i(t) \Phi\left(d_1(t)\right)-L \Phi\left(d_2(t)\right)\right], \\
d_1(t)=\frac{\ln \left(\frac{L_i(t)}{L}\right)+\frac{1}{2} \bar{\sigma}_i^2(t)}{\bar{\sigma}_i(t)}, d_2(t)=d_1(t)-\bar{\sigma}_i(t), \\
\bar{\sigma}_i^2(t)=\int_t^{t_{i-1}} \sigma^2(s) d s .
\end{aligned}
\end{align*}

(b) Today's price of a cap in the forward-LIBOR model $\operatorname{Cap}_{F L}(t ; V, L)$ with payment times $t_1<\ldots<t_N$ and level $L$ is given by,

$$
\operatorname{Cap}_{F L}(t ; V, L)=V \cdot \sum_{i=1}^N C_i\left(t, \sigma_i(t)\right) .
$$

In particular, if all volatility processes satisfy $\sigma_i(t)=\sigma$ for some positive constant $\sigma$ then we have

$$
\operatorname{Cap}_{F L}(t ; V, L)=\operatorname{Cap}_{\text {Black }}(t, V, L, \sigma),
$$
i.e. the price of the cap equals the one obtained with the Black formula.
\end{theorem}

\section*{Concepts of hedging.}

For hedging \footnote{\href{https://www.math.columbia.edu/~smirnov/over99.pdf}{Hedging concepts}} and risk management, the delta \cite{fliess2010delta},\cite{fliess2012preliminary} $\Delta=\frac{\partial V}{\partial S}$, i.e., the first partial derivative of the option's value with respect to the index level: For the Greeks of a European call option.

$$
\Delta=\frac{\partial C}{\partial S}=N\left(d_1\right)
$$

The gamma is the second partial derivative with respect to the index level,

$$
\gamma=\frac{\partial^2 C}{\partial S^2}=\frac{N^{\prime}\left(d_1\right)}{S \sigma \sqrt{T-t}}
$$

The theta of an option is, by convention, the negative first partial derivative with respect to time-to-maturity $t^*=T-t$

$$
\theta=-\frac{\partial C}{\partial t^*}=-\frac{S N^{\prime}\left(d_1\right) \sigma}{2 \sqrt{T-t}}-r e^{-r(T-t)} K N \left(d_2\right)
$$

The rho of an option is the first partial derivative with respect to the short rate $r$

$$
\rho=\frac{\partial C}{\partial r}=K(T-t) e^{-r(T-t)} N\left(d_2\right)
$$

The vega-which is obviously not a Greek letter-is the first partial derivative with respect to the volatility $\sigma$,

$$
\mathbf{vega}=\frac{\partial C}{\partial \sigma}=S N^{\prime}\left(d_1\right) \sqrt{T-t}
$$

Hedging dynamics can be defined as: 

$$
\Delta_t^P \equiv \frac{\partial P_t}{\partial S_t}
$$

$$
d P_t-\Delta_t^P d S_t=0
$$

Investment banks are also often interested in replicating the payoff of such a put (or another option). 

This is accomplished by setting up a replication portfolio consisting of $\Delta_t^P$ units of the underlying and $\gamma_t \equiv P_t-\Delta_t^P S_t$ units of the risk-less bond $B_t$ such that the resulting portfolio value equals the option value at any time $t$,

$$
P_t=\Delta_t^P S_t+\gamma_t B_t
$$
or
$$
d P_t=\Delta_t^P d S_t+\gamma_t d B_t
$$

For a plain vanilla put option this generally implies being short the underlying $\left(\Delta_t^P<0\right)$ and long the bond. For a call option $\left(\Delta_t^C>0\right)$ this implies the opposite.

A replication strategy $\left(\Delta_t^P, \gamma_t\right), t \in\{0, \ldots, \tau-\Delta t\}$, is called self-financing if for $t>0$ and $\tau$ being the exercise date (i.e, $\tau=T$ for a European option)
$$
\Delta_t^P S_t+\gamma_t B_t=\Delta_{t-\Delta t}^P S_t+\gamma_{t-\Delta t} B_t.
$$

Self-financing: The value $V_t$ of a portfolio containing $\psi_t$ units of bonds and $\phi_t=\left(\phi_{t, 1}, \ldots, \phi_{t, m}\right)^{\top}$ units of various stocks at a given time $t$ is given by
$$
V_t=B_t \psi_t+\sum_{i=1}^m \phi_{t, i} S_{t, i}=B_t \psi_t+\phi_t^{\top} \mathbf{S}_t .
$$

The particular value of the portfolio at time $t$ typically depends on the units of assets $\left(\psi_t, \phi_t\right)$ held just before time $t$. The stochastic process $\left\{\left(\psi_t, \phi_t\right), t \geqslant 0\right\}$ can thus be interpreted as a trading strategy.

 $\left\{\mathcal{F}_t, t \geqslant 0\right\}$. 
$$
\mathrm{d} V_t=\left(B_t \mathrm{~d} \psi_t+\mathbf{S}_t^{\top} \mathrm{d} \phi_t\right)+\psi_t \mathrm{~d} B_t+\phi_t^{\top} \mathrm{d} \mathbf{S}_t
$$

The portfolio is called self-financing if
$$
B_t \mathrm{~d} \psi_t+\mathbf{S}_t^{\top} \mathrm{d} \phi_t=0
$$
in which case
$$
V_T=V_0+\int_0^T \psi_t \mathrm{~d} B_{\ell}+\int_0^T \phi_t^{\top} \mathrm{d} \mathbf{S}_t .
$$

Arbitrage-free market: A market model is called arbitrage-free on $[0, T]$ if, in essence, for every self-financing trading strategy $\left\{\left(\psi_t, \phi_t\right), 0 \leqslant t \leqslant T\right\}$,
$$
V_0=0 \Rightarrow \mathbb{P}\left(V_T>0\right)=0 \text {. }
$$

 $\left\{\widetilde{\mathbf{S}}_t, 0 \leqslant t \leqslant T\right\}$, with $\widetilde{\mathbf{S}}_t=B_t^{-1} \mathbf{S}_t$, is an $\left(\mathcal{F}_t, \mathbb{Q}\right)$-martingale. 

$$
\sigma_t \mathbf{u}_t=\mu_t-r_t \mathbf{S}_t
$$
has a solution $\mathbf{u}_t, t \in[0, T]$ for which $\mathbb{E} \exp \left(\frac{1}{2} \int_0^T \mathbf{u}_t^{\top} \mathbf{u}_t \mathrm{~d} t\right)<\infty$

$$
\begin{aligned}
\mathrm{d} \widetilde{\mathbf{S}}_t & =B_t^{-1}\left(\boldsymbol{\mu}_t-r_t \mathbf{S}_t\right) \mathrm{d} t+B_t^{-1} \sigma_t \mathrm{~d} \mathbf{W}_t \\
& =B_t^{-1} \sigma_t\left(\mathbf{u}_t \mathrm{~d} t+\mathrm{d} \mathbf{W}_t\right) \\
& =B_t^{-1} \sigma_t \mathrm{~d} \mathbf{Z}_t
\end{aligned}
$$

where $\mathrm{d} \mathbf{Z}_t=\mathbf{u}_t \mathrm{~d} t+\mathrm{d} \mathbf{W}_t$ , defines an Itô process. Let

$$
M_t=\exp \left(\int_0^t \mathbf{u}_s^{\top} \mathrm{d} \mathbf{W}_s-\frac{1}{2} \int_0^t \mathbf{u}_s^{\top} \mathbf{u}_s \mathrm{~d} s\right) .
$$

Then, by Girsanov's theorem \cite{choe2016girsanov}, $\left\{M_t, t \geqslant 0\right\}$ is an $\left(\mathcal{F}_t, \mathbb{P}\right)$-martingale, and under the new measure $\mathbb{Q}(A) \stackrel{\text { def }}{=} \mathbb{E}\left[M_T \mathrm{I}_A\right]$ for all $A \in \mathcal{F}_T$, the process $\left\{\mathbf{Z}_t, 0 \leqslant\right.$ $t \leqslant T\}$ is a Wiener process.

It follows that under $\mathbb{Q}$ the process $\left\{\widetilde{\mathbf{S}}_t, 0 \leqslant t \leqslant T\right\}$ is a martingale with respect to $\left\{\mathcal{F}_t, 0 \leqslant t \leqslant T\right\}$, which had to be shown. $\mathbb{Q}$ is called the risk-neutral measure.

Let $\widetilde{V}_t=B_t^{-1} V_t=\psi_t+\boldsymbol{\phi}_t^{\top} B_t^{-1} \mathbf{S}_t=\psi_t+\boldsymbol{\phi}_t^{\top} \widetilde{\mathbf{S}}_t$ be the discounted value of the portfolio of stocks and bonds.

\begin{align*}
\begin{aligned}
\mathrm{d} \widetilde{V}_t=\boldsymbol{\phi}_t^{\top} \mathrm{d} \widetilde{\mathbf{S}}_t=B_t^{-1} \boldsymbol{\phi}_t^{\top} \boldsymbol{\sigma}_t \mathrm{~d} \mathbf{Z}_t
\end{aligned}
\end{align*}

$\mathbb{Q}$ the process $\left\{\tilde{V}_t, 0 \leqslant t \leqslant T\right\}$ is a martingale with respect to $\left\{\mathcal{F}_t, 0 \leqslant t \leqslant T\right\}$. As a consequence,
$$
V_t=B_t \mathbb{E}_{\mathbb{Q}}\left[B_T^{-1} V_T \mid \mathcal{F}_t\right], \quad t \leqslant T .
$$

The payoff at maturity for a call option is $\left(S_T-K\right)^{+}=\max \left\{S_T-K, 0\right\}$ and for a put option is $\left(K-S_T\right)^{+}$.

Suppose for concreteness that we are dealing with a call option and denote its value at time $t \leqslant T$ by $C_t=C\left(S_t, K, r_t, T-t\right)$, where $T-t$ is the time until maturity. 

 \begin{align*}
\begin{aligned}
C_t=V_t=B_t \mathbb{E}_{\mathbb{Q}}\left[B_T^{-1} C_T \mid \mathcal{F}_t\right], \quad t \leqslant T,
\end{aligned}
\end{align*}

where

\begin{align*}
\begin{aligned}
C_T=V_T=V_0+\int_0^T \psi_t \mathrm{~d} B_t+\int_0^T \boldsymbol{\phi}_t^{\top} \mathrm{d} \mathbf{S}_t .
\end{aligned}
\end{align*}

Consider a call and a put options written on the same non-dividend-paying stock, with the same strike price $K$ and time to maturity $T$. Then their prices $C_0$ and $P_0$ at time $t=0$ are related as follows:

$$
C_0+K e^{-r T}=P_0+S_0
$$

where $S_0$ is the current stock price and $r$ is the continuously compounded riskfree interest rate.

Consider a portfolio consisting of a call option and an amount of cash given by $K e^{-r T}$. At maturity, the value of this portfolio is,

$$
\max \left\{S_T-K, 0\right\}+K=\max \left\{S_T, K\right\}
$$

$$
\max \left\{K-S_T, 0\right\}+S_T=\max \left\{K, S_T\right\}
$$

Option prices provide information about state variables and parameters via the pricing equation. An option price is given by,

$$
C\left(S_t, V_t, \theta\right)=e^{-r(T-t)} \mathbb{E}_t^{\mathbb{Q}}\left[\max \left(S_t-K, 0\right) \mid S_t, X_t, \theta\right]
$$

where the expectation is taken under $\mathbb{Q}$ the risk-neutral probability measure. here we assume that the risk-free rate $r$ is constant. This simplifies by letting $A=\left\{S_T \geq K\right\}$ denote the event that the stock ends in the money. Then the option price is given by:

 \begin{align*}
\begin{aligned}
e^{-r t} \mathbb{E}_t^{\mathbb{Q}}\left(\max \left(S_T-K, 0\right)\right) & =e^{-r t} \mathbb{E}_t^{\mathbb{Q}}\left(S_T \mathbb{I}_A\right)-e^{-r t} K \mathbb{E}\left(\mathbb{I}_A\right) \\
& =e^{-r t} \mathbb{E}_t^{\mathbb{Q}}\left(S_T \mathbb{I}_A\right)-e^{-r t} K \mathbb{P}(A) .
\end{aligned}
\end{align*}

The logarithmic asset price and volatility follow an affine process with $Y_t=\left(\log S_t, V_t\right) \in \mathfrak{R} \times \mathfrak{R}^{+}$satisfying

\begin{align*}
\begin{aligned}
& d Y_t=\left(r-\frac{1}{2} V_t\right) d t+\sqrt{V_t} d B_{1, t} \\
& d V_t=\kappa\left(\theta-V_t\right) d t+\sigma_v \sqrt{V_t} d Z_t
\end{aligned}
\end{align*}

where $Z_t=\rho B_{1, t}+\sqrt{1-\rho^2} B_{2, t}$. The correlation, $\rho$, or so-called leverage effect is important to explain the empirical fact that volatility increases faster as equity prices drop. The parameters $(\kappa, \theta)$ govern the speed of mean reversion and the long-run mean of volatility and $\sigma_v$ measures the volatility of volatility. Under the risk-neutral measure they become $\left(\kappa / \kappa+\lambda_v\right) \theta$ and $\kappa+\lambda_v$ where $\lambda_v$ is the market price of volatility risk.

From affine process theory, the discounted transform,

$$
\psi(u)=e^{-r t} \mathbb{E}\left(e^{u Y_T} \mid Y_t\right)=e^{\alpha(t, u)+u Y_t+\beta(t, u) V_t}
$$
where $\alpha, \beta$ satisfy Riccati equations.

\section*{Concepts of Rough volatility.}

\begin{align*}
\begin{aligned}
W_t^H=C_H\left\{\int_{-\infty}^t \frac{d W_s^{\mathbb{P}}}{(t-s)^\gamma}-\int_{-\infty}^0 \frac{d W_s^{\mathbb{P}}}{(-s)^\gamma}\right\}
\end{aligned}
\end{align*}

\begin{align*}
\begin{aligned}
& \\
= & \nu C_H\left\{\int_t^u \frac{1}{(u-s)^\gamma} d W_s^{\mathbb{P}}+\int_{-\infty}^t\left[\frac{1}{(u-s)^\gamma}-\frac{1}{(t-s)^\gamma}\right] d W_s^{\mathbb{P}}\right\} \\
= & 2 \nu C_H\left[M_t(u)+Z_t(u)\right]
\end{aligned}
\end{align*}

$\mathbb{E}^{\mathbb{P}}\left[M_t(u) \mid \mathcal{F}_t\right]=0$ and $Z_t(u)$ is $\mathcal{F}_t$-measurable. 

Pricing \footnote{\href{https://tpq.io/p/rough_volatility_with_python.html}{Rough volatility: Python}} \footnote{\href{https://en.wikipedia.org/wiki/Rough_path}{Rough path}} \footnote{\href{https://hal.science/hal-03949577/document}{Rough volatility: Minimax Theory}} \cite{bondi2024rough},\cite{horvath2020volatility},\cite{mccrickerd2019spatially},\cite{keller2018affine}, under $\mathbb{P}$

$$
\tilde{W}_t^{\mathbb{P}}(u):=\sqrt{2 H} \int_t^u \frac{d W_s^{\mathrm{P}}}{(u-s)^\gamma}
$$

With $\eta:=2 \nu C_H / \sqrt{2 H}$ we have $2 \nu C_H M_t(u)=\eta \tilde{W}_t^{\mathbb{P}}(u)$ ,

\begin{align*}
\begin{aligned}
v_u & =v_t \exp \left\{\eta \tilde{W}_t^{\mathbb{P}}(u)+2 \nu C_H Z_t(u)\right\} \\
& =\mathbb{E}^{\mathbb{P}}\left[v_u \mid \mathcal{F}_t\right] \mathcal{E}\left(\eta \tilde{W}_t^{\mathbb{P}}(u)\right)
\end{aligned}
\end{align*}

\begin{align*}
\begin{aligned}
v_u=\mathbb{E}^{\mathbb{P}}\left[v_u \mid \mathcal{F}_t\right] \mathcal{E}\left(\eta \tilde{W}_t^{\mathbb{P}}(u)\right)
\end{aligned}
\end{align*}

$$
d W_s^{\mathrm{P}}=d W_s^{\mathbb{Q}}+\lambda_s d s
$$

where $\left\{\lambda_s: s>t\right\}$ has a natural interpretation as the price of volatility risk.

\begin{align*}
\begin{aligned}
v_u=\mathbb{E}^{\mathbb{P}}\left[v_u \mid \mathcal{F}_t\right] \mathcal{E}\left(\eta \tilde{W}_t^Q(u)\right) \exp \left\{\eta \sqrt{2 H} \int_t^u \frac{\lambda_s}{(u-s)^\gamma} d s\right\}
\end{aligned}
\end{align*}

\section*{Concepts of Fourier-based option pricing.}

\subsection{Bayesian optimization for option pricing based on the Fourier method.}

In this section we apply a Fourier Transform to calculate the option price using a bayesian optimization.
The results are available on ART.1.20 \footnote{\href{https://github.com/Trusted-AI/adversarial-robustness-toolbox/pull/2467}{Rough volatility: Python}}

\begin{algorithm}
\small
\caption{Fourier Transform Method for European Call Option Price }
\label{alg:fourier_transform_option_price}
\begin{algorithmic}[1]
\Procedure{FourierTransformOptionPrice}{$S,K,T,r,\sigma,\alpha,\beta$}
    \State $u \gets \text{linspace}(-10,10,1000)$ 
    \State $C \gets [\text{characteristic\_function}(u_i) \mid u_i \in u]$ 
    \State $\text{option\_price} \gets \text{inverse\_fourier\_transform}(u,C)$ 
    \State \Return $\text{option\_price}$
\EndProcedure
\end{algorithmic}
\end{algorithm}

\begin{align*}
\begin{aligned}
\langle f, g\rangle & =\int_{-\infty}^{\infty} \frac{1}{2 \pi} \int_{-\infty}^{\infty} e^{-i u x} \hat{f}(k) d k \overline{g(x)} d x \\
& =\frac{1}{2 \pi} \int_{-\infty}^{\infty} \hat{f}(k) \int_{-\infty}^{\infty} e^{-i u x} \overline{g(x)} d x d k \\
& =\frac{1}{2 \pi} \int_{-\infty}^{\infty} \hat{f}(k) \int_{-\infty}^{\infty} \overline{e^{i u x} g(x)} d x d k \\
& =\frac{1}{2 \pi} \int_{-\infty}^{\infty} \hat{f}(k) \overline{\hat{g}(k)} d k
\end{aligned}
\end{align*}

\begin{align*}
\begin{aligned}
\langle f, g\rangle \equiv \int_{-\infty}^{\infty} f(x) \overline{g(x)} d x
\end{aligned}
\end{align*}. 

By Fourier inversion $f(x)=$ $\frac{1}{2 \pi} \int_{-\infty}^{\infty} e^{-i u x} \hat{f}(k) d k$.

Let a random variable $X$ be distributed with pdf $q(x)$. 

The characteristic function $\hat{q}$ of $X$ is the Fourier transform \cite{sakurai2024learning},\cite{zhang2013fast},\cite{wong2011fft},\cite{wang2022pricing} of its $p d f$,

$$
\hat{q}(u) \equiv \int_{-\infty}^{\infty} e^{i u x} q(x) d x=\mathbf{E}^Q\left(e^{i u X}\right)
$$

Consider a European call option with payoff 
$C_T \equiv \max \left[e^s-K, 0\right]$ where $s \equiv \log S$.

Call Option Transform \cite{hurd2010fourier},\cite{abi2024signature}:
For $u=u_r+i u_i$ with $u_i>1$, 

the Fourier \footnote{\href{https://www.math.hkust.edu.hk/~maykwok/piblications/Handbook_chapter.pdf}{Pricing: Fast Fourier
Transform}} \footnote{\href{https://kristofkassa.github.io/derivative_pricing/2023/10/17/fourier_methods.html}{derivative pricing}} \footnote{\href{https://www.intechopen.com/chapters/36433}{Fourier Transform Applications}}transform of $C_T$ is,

$$
\widehat{C}_T(u)=-\frac{K^{i u+1}}{u^2-i u}
$$

\begin{proof}

\begin{align*}
\begin{aligned}
\widehat{C}_T(u) & =\int_{-\infty}^{\infty} e^{i u s} \max \left[e^s-K, 0\right] d s \\
& =\int_{\log K}^{\infty} e^{i u s}\left(e^s-K\right) d s
\end{aligned}
\end{align*}

\begin{align*}
\begin{aligned}
& =\int_{\log K}^{\infty}\left(e^{(i u+1) s}-K e^{i u s}\right) d s \\
& =\left[\frac{e^{(i u+1) s}}{i u+1}-K \frac{e^{i u s}}{i u}\right]_{\log K}^{\infty} \\
& =-\frac{K^{i u+1}}{u^2-i u}
\end{aligned}
\end{align*}

\begin{align*}
\begin{aligned}
C_T(s)=\frac{1}{2 \pi} \int_{-\infty+i u_i}^{\infty+i u_i} e^{-i u s} \widehat{C}_T(u) d u
\end{aligned}
\end{align*}

\begin{align*}
\begin{aligned}
C_0 & =e^{-r T} \mathbf{E}_0^Q\left(C_T\right) \\
& =\frac{e^{-r T}}{2 \pi} \mathbf{E}_0^Q\left(\int_{-\infty}^{\infty+i u_i} e^{-i u s} \widehat{C}_T(u) d u\right) \\
& =\frac{e^{-r T}}{2 \pi} \int_{-\infty+i u_i}^{\infty+i u_i} \mathbf{E}_0^Q\left(e^{i(-u) s}\right) \widehat{C}_T(u) d u \\
& =\frac{e^{-r T}}{2 \pi} \int_{-\infty+i u_i}^{\infty+i u_i} \widehat{C}_T(u) \hat{q}(-u) d u
\end{aligned}
\end{align*}

If $S_t \equiv S_0 e^{r t+X_t}$ with $X_t$ a Lévy process and $e^{X_t}$ a martingale with $X_0=0$, 

then $\hat{q}(-u)=$ $e^{-i u y} \varphi(-u)$ where $\varphi$ is the characteristic function of $X_T$. Here, $y \equiv \log S_0+r T$. 

$$
C_0=\frac{e^{-r T}}{2 \pi} \int_{-\infty+i u_i}^{\infty+i u_i} e^{-i u y} \widehat{C}(u) \varphi(-u) du
$$

$k=\log \left(S_0 / K\right)+r T$,

$$
C_0=-\frac{K e^{-r T}}{2 \pi} \int_{-\infty+i u_i}^{\infty+i u_i} e^{-i u k} \varphi(-u) \frac{d u}{u^2-u i}
$$

$u_i \in(0,1)$, the call option present value is,

$$
C_0=S_0-\frac{K e^{-r T}}{2 \pi} \int_{-\infty+i u_i}^{\infty+i u_i} e^{-i u k} \varphi(-u) \frac{d u}{u^2-u i}
$$

$u_i=0.5$ ,

$$
C_0=S_0-\frac{\sqrt{S_0 K} e^{-r T / 2}}{\pi} \int_0^{\infty} \operatorname{Re}\left[e^{i z k} \varphi(z-i / 2)\right] \frac{d z}{z^2+1 / 4}
$$

where $\mathbf{R e}[x]$ denotes the real part of $x$. $u=i$ ,  $u=0$, 

$$
\begin{aligned}
\operatorname{Res}(i) & =\lim _{u \rightarrow i}\left((u-i)\left(-\frac{K e^{-r T}}{2 \pi} e^{-i u k} \frac{\varphi(-u)}{u(u-i)}\right)\right) \\
& =-\frac{K e^{-r T}}{2 \pi} e^k \frac{\varphi(-i)}{i} \\
& =\frac{S_0 i}{2 \pi}
\end{aligned}
$$

$e^k=S_0 / K \cdot e^{r T}, \varphi(-i)=1$ 

and $i^{-1}=-i$ in the  $u_i$ minus $2 \pi i \operatorname{Res}(i)=-S_0$, $u_i=0.5$

\begin{align*}
\begin{aligned}
C_0 = S_0-\frac{K e^{-r T}}{2 \pi}\int_{-\infty}^{\infty} \\ e^{-i(u+i / 2) k} \varphi(-(u+i / 2)) \frac{d u}{(u+i / 2)^2-(u+i / 2) i}
\end{aligned}
\end{align*}

$$
e^{-i(u+i / 2) k}=e^{-i u k} e^{k / 2}
$$

$$
e^{k / 2}=e^{\left(\log \left(S_0 / K\right)+r T\right) / 2}
$$

$$
K e^{-r T} e^{k / 2}=e^{-r T / 2} \sqrt{S_0 K}
$$

$u=z+i / 2$ , 

$$
(z-i / 2)^2-(z-i / 2) i=z^2-2 z i+1 / 4
$$

$$
C_0=S_0-\frac{\sqrt{S_0 K} e^{-r T / 2}}{\pi} \int_{-\infty}^{\infty} e^{-i z k} \varphi(-z-i / 2) \frac{d z}{z^2-2 z i+1 / 4}
$$

\begin{align*}
\begin{aligned}
f(x) & =\frac{1}{2 \pi} \int_{-\infty}^{\infty} e^{-i u x} \hat{f}(u) d u \\
& =\frac{1}{2 \pi} \operatorname{Re}\left[\int_{-\infty}^0 e^{-i u x} \hat{f}(u) d u\right]+\frac{1}{2 \pi} \int_0^{\infty} \operatorname{Re}\left[e^{-i u x} \hat{f}(u) d u\right] \\
& =\frac{1}{2 \pi} \operatorname{Re}\left[\int_0^{\infty} e^{-i u x} \hat{f}(u) d u\right]+\frac{1}{2 \pi} \int_0^{\infty} \operatorname{Re}\left[e^{-i u x} \hat{f}(u) d u\right]
\end{aligned}
\end{align*}

\begin{align*}
\begin{aligned}
& =\frac{1}{2 \pi} \mathbf{R e}\left[2 \int_0^{\infty} e^{-i u x} \hat{f}(u) d u\right] \\
& =\frac{1}{\pi} \mathbf{R e}\left[\int_0^{\infty} e^{-i u x} \hat{f}(u) d u\right]
\end{aligned}
\end{align*}

$$
C_0=S_0-\frac{\sqrt{S_0 K} e^{-r T / 2}}{\pi} \int_0^{\infty} \operatorname{Re}\left[e^{-i z k} \varphi(-z-i / 2)\right] \frac{d z}{z^2+1 / 4}
$$

$C_T \equiv$ $\max \left[S_T-K, 0\right]$ , $K \equiv e^k$ and $S_T \equiv e^s$ ,
 
\begin{align*}
\begin{aligned}
C_0 & \equiv e^{-r T} \mathbf{E}_0^Q\left(\max \left[e^s-e^k, 0\right]\right) \\
& =e^{-r T} \int_k^{\infty}\left(e^s-e^k\right) q(s) d s
\end{aligned}
\end{align*}

$q(s)$ is the risk-neutral pdf of $s_T$, $c_0 \equiv e^{\alpha k} C_0$ with $\alpha>0$. 

The Fourier transform of $c_0$ is

$$
\psi(v) \equiv \int_{-\infty}^{\infty} e^{i v k} c_0 d k
$$

$$
C_0=\frac{e^{-\alpha k}}{\pi} \int_0^{\infty} e^{-i v k} \psi(v) d v
.$$

\end{proof}

\section*{Concepts of Risk.} 

\begin{figure}
\centering
\includegraphics[width=3.2in]{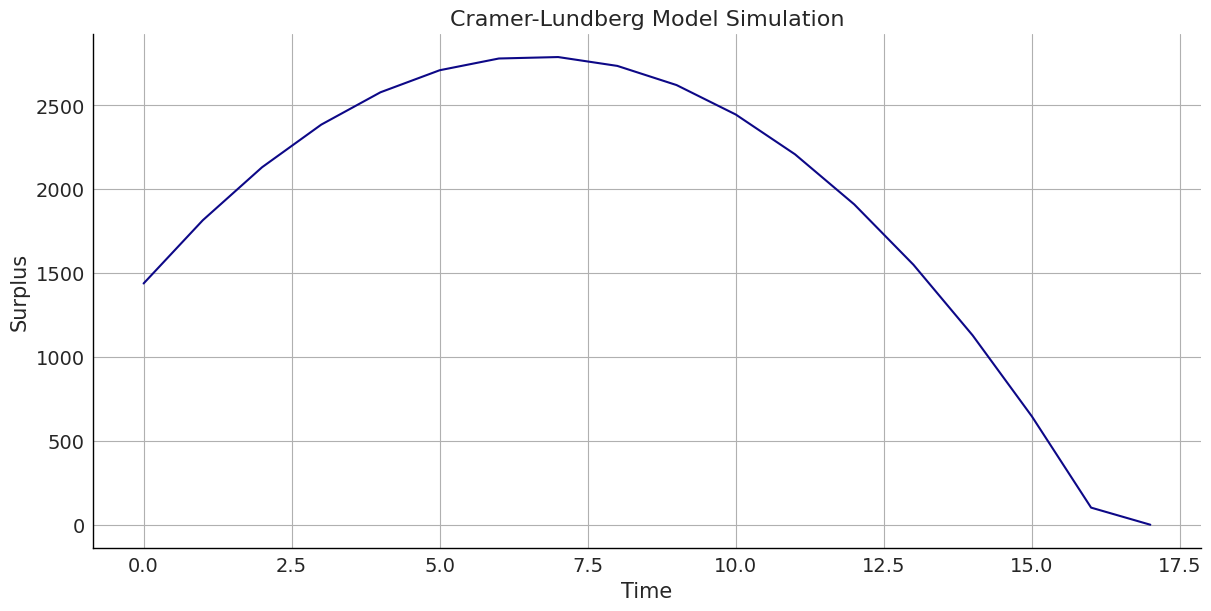}
\caption{Cramer-Lundberg model via bayesian optimization}
\label{fig:model_cramer_lundberg}
\end{figure}

To quantify the measurement of financial risk, we apply ‘Cramer-Lundberg model (Figure \ref{fig:model_cramer_lundberg})’ \cite{aurzada2020ruin},\cite{braunsteins2023cramer},\cite{xiong2011ruin},\cite{weber2024critical}, ‘VaR’, ‘Tail probability’ and ‘shortfall integral’ methods via a bayesian approach, 
the results obtained can be viewed on: {\color{blue} \url{https://github.com/Trusted-AI/adversarial-robustness-toolbox/pull/2467}}.

Let us consider a random variable, $X$, with density function, $f_X(x)$. The moment-generating function is defined as,

\begin{align*}
\begin{aligned}
M_X(t) & =\mathbb{E}\left(e^{t X}\right) \\
& =\int_{-\infty}^{\infty} e^{t x} f_X(x) d x
\end{aligned}
\end{align*}

$$
f_X(x)=\frac{1}{2 \pi i} \int_{-i \infty,(0+)}^{+i \infty} M_X(t) e^{-t x} d t
$$

for $x \in \mathbb{R}, t \in \mathbb{C}$ and where $i$ is the imaginary number satisfying $i=\sqrt{-1}$, we introduce $K_X(t)=\ln M_X(t)$,

\begin{align*}
\begin{aligned}
f_X(x) & =\frac{1}{2 \pi i} \int_{-i \infty,(0+)}^{+i \infty} e^{\ln M_X(t)} e^{-t x} d t \\
& =\frac{1}{2 \pi i} \int_{-i \infty,(0+)}^{+i \infty} e^{K_X(t)-t x} dt  ({\text  {Equation }  a})
\end{aligned}
\end{align*}

\section*{The Tail Probability}

$$
\mathbb{P}(X>x)=\int_x^{\infty} f_X(u) d u
$$

\begin{align*}
\begin{aligned}
\mathbb{P}(X>x) & =\int_x^{\infty} \underbrace{\frac{1}{2 \pi i} \int_{-i \infty,(0+)}^{+i \infty} e^{K_X(t)-t u} d t}_{\text {Equation } a} d u \\
& =\frac{1}{2 \pi i} \int_{-i \infty,(0+)}^{+i \infty} \int_x^{\infty} e^{K_X(t)-t u} d u d t \\
& =\frac{1}{2 \pi i} \int_{-i \infty,(0+)}^{+i \infty} e^{K_X(t)}\left(\int_x^{\infty} e^{-t u} d u\right) d t \\
& =\frac{1}{2 \pi i} \int_{-i \infty,(0+)}^{+i \infty} e^{K_X(t)}\left[-\frac{e^{-t u}}{t}\right]_x^{\infty} d t \\
& =\frac{1}{2 \pi i} \int_{-i \infty,(0+)}^{+i \infty} e^{K_X(t)}\left[0-\left(-\frac{e^{-t x}}{t}\right)\right] d t \\
& =\frac{1}{2 \pi i} \int_{-i \infty,(0+)}^{+i \infty} \frac{e^{K_X(t)-t x}}{t} d t
\end{aligned}
\end{align*}

The VaR\footnote{\href{https://en.wikipedia.org/wiki/Expected_shortfall}{VaR}} is an upper-tail probability, which allows us to deduce:

\section*{shortfall integral}

\begin{align*}
\begin{aligned}
& \mathbb{E}(X \mid X>x)=\frac{1}{\mathbb{P}(X>x)} \int_x^{\infty} u f_X(u) d u, \\
& =\frac{1}{\mathbb{P}(X>x)} \int_x^{\infty} u \underbrace{\frac{1}{2 \pi i} \int_{-i \infty,(0+)}^{+i \infty} e^{K_X(t)-t u} d t}_{\text {Equation } a} d u, \\
& =\frac{1}{\mathbb{P}(X>x)} \frac{1}{2 \pi i} \int_{-i \infty,(0+)}^{+i \infty} e^{K_X(t)}\left(\int_x^{\infty} u e^{-t u} d u\right) d t, \\
& =\frac{1}{\mathbb{P}(X>x)} \frac{1}{2 \pi i} \int_{-i \infty,(0+)}^{+i \infty} e^{K_X(t)}\left[\left(-\frac{u}{t}-\frac{1}{t^2}\right) e^{-t u}\right]_x^{\infty} d t, \\
& =\frac{1}{\mathbb{P}(X>x)} \frac{1}{2 \pi i} \int_{-i \infty,(0+)}^{+i \infty} e^{K_X(t)}\left[-\left(-\frac{x}{t}-\frac{1}{t^2}\right) e^{-t x}\right] d t \text {, } \\
& =\frac{1}{\mathbb{P}(X>x)} \frac{1}{2 \pi i} \int_{-i \infty,(0+)}^{+i \infty} e^{K_X(t)}\left[\frac{x e^{-t x}}{t}+\frac{e^{-t x}}{t^2}\right] d t \text {, } \\
& =\frac{1}{\mathbb{P}(X>x)}(\frac{1}{2 \pi i} \int_{-i \infty,(0+)}^{+i \infty} \frac{e^{K_X(t)-t x}}{t^2} d t+x \underbrace{\frac{1}{2 \pi i} \int_{-i \infty,(0+)}^{+i \infty} \frac{e^{K_X(t)-t x}}{t} d t}_{\mathbb{P}(X>x)}) \text {, } \\
& =\frac{1}{\mathbb{P}(X>x)}\left(\frac{1}{2 \pi i} \int_{-i \infty,(0+)}^{+i \infty} \frac{e^{K_X(t)-t x}}{t^2} d t+x \mathbb{P}(X>x)\right) \text {, } \\
& =x+\frac{1}{\mathbb{P}(X>x)}\left(\frac{1}{2 \pi i} \int_{-i \infty,(0+)}^{+i \infty} \frac{e^{K_X(t)-t x}}{t^2} d t\right) \text {. }
\end{aligned}
\end{align*}

\section*{Lundberg's inequality}

\begin{theorem}

Suppose $\kappa>0$. Then the probability of ruin $\psi(u)$ satisfies,

$$
\psi(u) \leq e^{-\kappa u}, \quad u \geq 0
$$

\end{theorem}

\begin{proof}

\begin{align*}
\begin{aligned}
\psi_{n+1}(u) \\ = \int_0^{\infty}\left[1-F(u+c t)+ \int_0^{u+c t} \psi_n(u+c t-x) d F(x)\right] \lambda e^{-\lambda t} dt 
\end{aligned}
\end{align*}

\begin{align*}
\begin{aligned}
 & \hspace{-1cm} \psi_{n+1}(u)= \int_0^{\infty}\left[\int_{u+c t}^{\infty} d F(x)+\int_0^{u+c t} \psi_n(u+c t-x) d F(x)\right] \lambda e^{-\lambda t} d t \\
& \leq \int_0^{\infty}\left[\int_{u+c t}^{\infty} e^{-\kappa(u+c t-x)} d F(x)\right. \\
&\left.+\int_0^{u+c t} e^{-\kappa(u+c t-x)} d F(x)\right] \lambda e^{-\lambda t} d t
\end{aligned}
\end{align*}

\begin{align*}
\begin{aligned}
\psi_{n+1}(u) & \leq \int_0^{\infty}\left[\int_0^{\infty} e^{-\kappa(u+c t-x)} d F(x)\right] \lambda e^{-\lambda t} d t \\
& =\lambda e^{-\kappa u} \int_0^{\infty} e^{-\kappa c t}\left[\int_0^{\infty} e^{\kappa x} d F(x)\right] e^{-\lambda t} d t \\
& =\lambda e^{-\kappa u} \int_0^{\infty} e^{-(\lambda+\kappa c) t}\left[M_X(\kappa)\right] d t \\
& =\lambda M_X(\kappa) e^{-\kappa u} \int_0^{\infty} e^{-(\lambda+\kappa c) t} d t \\
& =\frac{\lambda M_X(\kappa)}{\lambda+\kappa c} e^{-\kappa u}
\end{aligned}
\end{align*}

$$
\lambda M_X(\kappa)=\lambda[1+(1+\theta) \kappa \mu]=\lambda+\kappa(1+\theta) \lambda \mu=\lambda+\kappa c
$$

$\psi_{n+1}(u) \leq e^{-\kappa u}$ , $\psi_n(u) \leq e^{-\kappa u}$ for all $n$ , $\psi(u)=$ $\lim _{n \rightarrow \infty} \psi_n(u) \leq e^{-\kappa u}$.

$$
\theta=\frac{u\left\{\mathrm{E}\left[\exp \left(-\frac{\ln \alpha}{u} X\right)\right]-1\right\}}{-\mu \ln \alpha}-1
$$

$\kappa=(-\ln \alpha) / u$, $\psi(u) \leq e^{-\kappa u}=e^{\ln \alpha}=\alpha$,

$$
u=\frac{-\ln \alpha}{\kappa}
$$

 $\psi(u) \leq e^{-\kappa u}=e^{\ln \alpha}=\alpha$ 

$$
\psi(\infty)=\lim _{u \rightarrow \infty} \psi(u)=0
$$

$$
0 \leq \psi(u) \leq e^{-\kappa u}
$$

\hspace{1.5cm} Suppose $\kappa>0$. Then the ruin probability satisfies,

$$
\psi(u) \sim C e^{-\kappa u}, \quad u \rightarrow \infty
$$

$$
C=\frac{\mu \theta}{M_X^{\prime}(\kappa)-\mu(1+\theta)}
$$

\end{proof}

\section*{Understanding Diffusion concepts at a more advanced level }

\section*{Concepts of Diffusion Riemannian.}

Consider a solution $u: \mathcal{M} \times[0, T] \rightarrow[0, \infty)$ ,

$$
\frac{\partial u}{\partial t}=\Delta u
$$

with $\int_{\mathcal{M}} u(\cdot, 0) d \mu_g=1$, where $\mu_g$ is the Riemannian \footnote{\href{https://bjlkeng.io/posts/hyperbolic-geometry-and-poincare-embeddings/}{Ricci-Riemann}} volume measure.

$$
\frac{d}{d t} \int_{\mathcal{M}} u(\cdot, t) d \mu_g=\int_{\mathcal{M}} \Delta u d \mu_g=0,
$$

The integral of $u$ over $\mathcal{M}$ remains one for each $t \in[0, T]$, and in particular, $u$ can be viewed as a probability density of a measure $\nu(t)$ defined by $d \nu(t):=u(\cdot, t) d \mu_g$.

$\nu(t)$ represents the probability distribution of a particle moving on the manifold under Brownian motion, with initial probability distribution $\nu(0)$,

Suppose $g(\tau)$ is a family of Riemannian metrics on $\mathcal{M}$, for $\tau \in$ $[0, T]$. We call a flow of measures $v(\tau)$ a diffusion \cite{chen2023geometric} (representing the probability distribution of a Brownian particle as above) if $d \nu(\tau)=u(\cdot, \tau) d \mu_{g(\tau)}$ with , 

$$
\frac{\partial u}{\partial \tau}=\Delta_{g(\tau)} u-\left(\frac{1}{2} \operatorname{tr} \frac{\partial g}{\partial \tau}\right) u
$$

$\frac{\partial g}{\partial \tau}=2 \operatorname{Ric}(g(\tau))$, $g(\cdot)$ a Ricci flow with respect to a 'reverse' time coordinate $\tau$ (that is, $\tau=C-t$ for some constant $C$ ),

$$
\frac{\partial u}{\partial \tau}=\Delta_{g(\tau)} u-R u
$$

$$
\begin{aligned} 
\frac{d}{d \tau} \int_{\mathcal{M}} d \nu(\tau) \\ = \frac{d}{d \tau} \int_{\mathcal{M}} u(\cdot, \tau) d \mu_{g(\tau)} & =\\\int_{\mathcal{M}}\left(\frac{\partial u}{\partial \tau} d \mu_{g(\tau)}+u \frac{\partial}{\partial \tau} d \mu_{g(\tau)}\right) \\
& =\int_{\mathcal{M}} \Delta_{g(\tau)} u d \mu_{g(\tau)}=0
\end{aligned}
$$

 If $g(\tau)$ is a flow of Riemannian metrics for $\tau \in[a, b]$, and $\nu(\tau)$ is a diffusion (as defined above) then given any $f: \mathcal{M} \times[a, b] \rightarrow \mathbb{R}$ solving $-\frac{\partial f}{\partial \tau}=\Delta_{g(\tau)} f$, 

$$
\frac{d}{d \tau} \int_{\mathcal{M}} f(\cdot, \tau) d \nu(\tau)=0 .
$$

\section*{Concepts of McKean-Vlasov process.}

\begin{figure}[H] 
\centering
\includegraphics[width=3.3in]{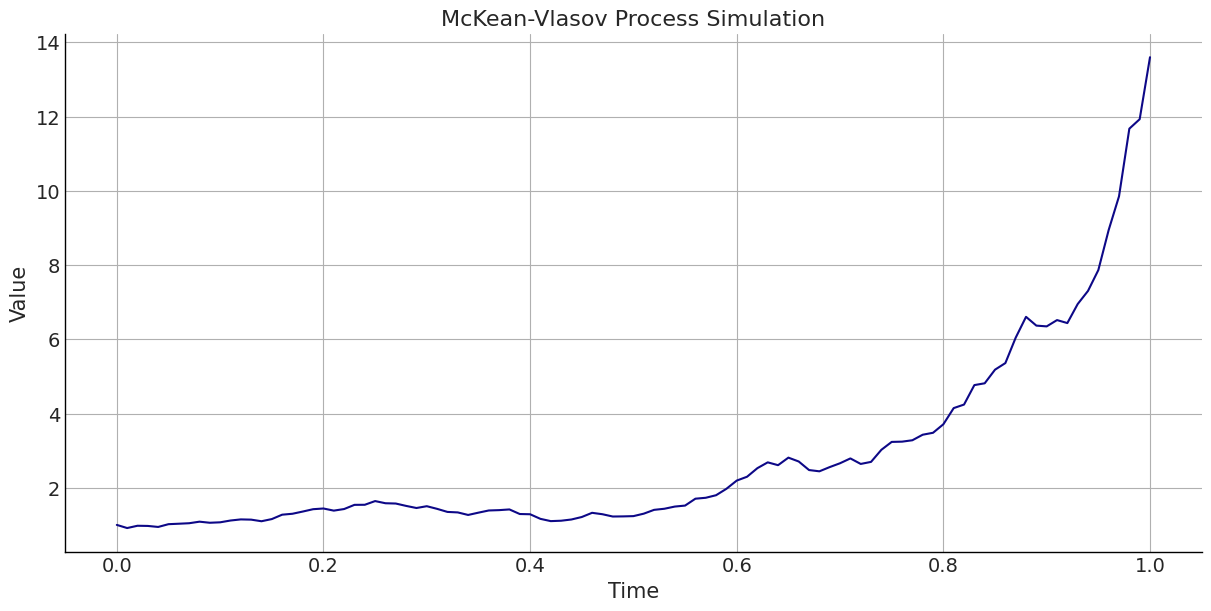}
\caption{McKean-Vlasov process via bayesian optimization}
\label{fig:McKean-Vlasov process}
\end{figure}

This conceptualization of diffusion models in Riemman's sense allows us to state “Propagation of Chaos”, through a “McKean-Vlasov process (Figure \ref{fig:McKean-Vlasov process})” \cite{yang2024neural}, \cite{coghi2018pathwise}, \cite{carmona2013control}, \cite{han2024learning}, \cite{qian2022mckean},\cite{adams2022large},\cite{meherrem2019maximum} incorporating Bayesian optimization of the data distribution. The results obtained can be viewed on ART.1.18\footnote{\href{https://github.com/Trusted-AI/adversarial-robustness-toolbox/pull/2467}{ART-IBM}}

\bibliographystyle{IEEEtran}

\bibliography{IEEEabrv,refs}

\begin{thebibliography}{100}
\providecommand{\url}[1]{#1}
\csname url@samestyle\endcsname
\providecommand{\newblock}{\relax}
\providecommand{\bibinfo}[2]{#2}
\providecommand{\BIBentrySTDinterwordspacing}{\spaceskip=0pt\relax}
\providecommand{\BIBentryALTinterwordstretchfactor}{4}
\providecommand{\BIBentryALTinterwordspacing}{\spaceskip=\fontdimen2\font plus
\BIBentryALTinterwordstretchfactor\fontdimen3\font minus \fontdimen4\font\relax}
\providecommand{\BIBforeignlanguage}[2]{{%
\expandafter\ifx\csname l@#1\endcsname\relax
\typeout{** WARNING: IEEEtran.bst: No hyphenation pattern has been}%
\typeout{** loaded for the language `#1'. Using the pattern for}%
\typeout{** the default language instead.}%
\else
\language=\csname l@#1\endcsname
\fi
#2}}
\providecommand{\BIBdecl}{\relax}
\BIBdecl

\bibitem{yenduri2024gpt}
G.~Yenduri, M.~Ramalingam, G.~C. Selvi, Y.~Supriya, G.~Srivastava, P.~K.~R. Maddikunta, G.~D. Raj, R.~H. Jhaveri, B.~Prabadevi, W.~Wang \emph{et~al.}, ``Gpt (generative pre-trained transformer)--a comprehensive review on enabling technologies, potential applications, emerging challenges, and future directions,'' \emph{IEEE Access}, 2024.

\bibitem{gupta2023chatgpt}
M.~Gupta, C.~Akiri, K.~Aryal, E.~Parker, and L.~Praharaj, ``From chatgpt to threatgpt: Impact of generative ai in cybersecurity and privacy,'' \emph{IEEE Access}, 2023.

\bibitem{troxler2024actuarial}
A.~Troxler and J.~Schelldorfer, ``Actuarial applications of natural language processing using transformers: Case studies for using text features in an actuarial context,'' \emph{British Actuarial Journal}, vol.~29, p.~e4, 2024.

\bibitem{yao2024survey}
Y.~Yao, J.~Duan, K.~Xu, Y.~Cai, Z.~Sun, and Y.~Zhang, ``A survey on large language model (llm) security and privacy: The good, the bad, and the ugly,'' \emph{High-Confidence Computing}, p. 100211, 2024.

\bibitem{wang2021blockchain}
Z.~Wang and Q.~Hu, ``Blockchain-based federated learning: A comprehensive survey,'' \emph{arXiv preprint arXiv:2110.02182}, 2021.

\bibitem{sun2024trustllm}
L.~Sun, Y.~Huang, H.~Wang, S.~Wu, Q.~Zhang, C.~Gao, Y.~Huang, W.~Lyu, Y.~Zhang, X.~Li \emph{et~al.}, ``Trustllm: Trustworthiness in large language models,'' \emph{arXiv preprint arXiv:2401.05561}, 2024.

\bibitem{cao2022ai}
L.~Cao, ``Ai in finance: challenges, techniques, and opportunities,'' \emph{ACM Computing Surveys (CSUR)}, vol.~55, no.~3, pp. 1--38, 2022.

\bibitem{wu2022sustainable}
C.-J. Wu, R.~Raghavendra, U.~Gupta, B.~Acun, N.~Ardalani, K.~Maeng, G.~Chang, F.~Aga, J.~Huang, C.~Bai \emph{et~al.}, ``Sustainable ai: Environmental implications, challenges and opportunities,'' \emph{Proceedings of Machine Learning and Systems}, vol.~4, pp. 795--813, 2022.

\bibitem{zawish2024ai}
M.~Zawish, F.~A. Dharejo, S.~A. Khowaja, S.~Raza, S.~Davy, K.~Dev, and P.~Bellavista, ``Ai and 6g into the metaverse: Fundamentals, challenges and future research trends,'' \emph{IEEE Open Journal of the Communications Society}, vol.~5, pp. 730--778, 2024.

\bibitem{bommasani2021opportunities}
R.~Bommasani, D.~A. Hudson, E.~Adeli, R.~Altman, S.~Arora, S.~von Arx, M.~S. Bernstein, J.~Bohg, A.~Bosselut, E.~Brunskill \emph{et~al.}, ``On the opportunities and risks of foundation models,'' \emph{arXiv preprint arXiv:2108.07258}, 2021.

\bibitem{pan2023large}
J.~Z. Pan, S.~Razniewski, J.-C. Kalo, S.~Singhania, J.~Chen, S.~Dietze, H.~Jabeen, J.~Omeliyanenko, W.~Zhang, M.~Lissandrini \emph{et~al.}, ``Large language models and knowledge graphs: Opportunities and challenges,'' \emph{arXiv preprint arXiv:2308.06374}, 2023.

\bibitem{mengara2024art}
O.~Mengara, ``The art of deception: Robust backdoor attack using dynamic stacking of triggers,'' \emph{arXiv preprint arXiv:2401.01537}, 2024.

\bibitem{li2023large}
Y.~Li, S.~Wang, H.~Ding, and H.~Chen, ``Large language models in finance: A survey,'' in \emph{Proceedings of the fourth ACM international conference on AI in finance}, 2023, pp. 374--382.

\bibitem{peng2021survey}
K.~Peng and G.~Yan, ``A survey on deep learning for financial risk prediction,'' \emph{Quantitative Finance and Economics}, vol.~5, no.~4, pp. 716--737, 2021.

\bibitem{lee2024survey}
J.~Lee, N.~Stevens, S.~C. Han, and M.~Song, ``A survey of large language models in finance (finllms),'' \emph{arXiv preprint arXiv:2402.02315}, 2024.

\bibitem{nie2024survey}
Y.~Nie, Y.~Kong, X.~Dong, J.~M. Mulvey, H.~V. Poor, Q.~Wen, and S.~Zohren, ``A survey of large language models for financial applications: Progress, prospects and challenges,'' \emph{arXiv preprint arXiv:2406.11903}, 2024.

\bibitem{huang2024survey}
X.~Huang, W.~Ruan, W.~Huang, G.~Jin, Y.~Dong, C.~Wu, S.~Bensalem, R.~Mu, Y.~Qi, X.~Zhao \emph{et~al.}, ``A survey of safety and trustworthiness of large language models through the lens of verification and validation,'' \emph{Artificial Intelligence Review}, vol.~57, no.~7, p. 175, 2024.

\bibitem{mengara2024last}
O.~Mengara, ``The last dance: Robust backdoor attack via diffusion models and bayesian approach,'' \emph{arXiv preprint arXiv:2402.05967}, 2024.

\bibitem{mengara2024trading}
------, ``Trading devil: Robust backdoor attack via stochastic investment models and bayesian approach,'' arXiv. org, Tech. Rep., 2024.

\bibitem{joshi2008new}
M.~Joshi and A.~Stacey, ``New and robust drift approximations for the libor market model,'' \emph{Quantitative Finance}, vol.~8, no.~4, pp. 427--434, 2008.

\bibitem{bayer2023rough}
C.~Bayer, P.~K. Friz, M.~Fukasawa, J.~Gatheral, A.~Jacquier, and M.~Rosenbaum, \emph{Rough volatility}.\hskip 1em plus 0.5em minus 0.4em\relax SIAM, 2023.

\bibitem{gatheral2018volatility}
J.~Gatheral, T.~Jaisson, and M.~Rosenbaum, ``Volatility is rough,'' \emph{Quantitative finance}, vol.~18, no.~6, pp. 933--949, 2018.

\bibitem{fukasawa2024partial}
M.~Fukasawa and R.~Takano, ``A partial rough path space for rough volatility,'' \emph{Electronic Journal of Probability}, vol.~29, pp. 1--28, 2024.

\bibitem{abi2019lifting}
E.~Abi~Jaber, ``Lifting the heston model,'' \emph{Quantitative finance}, vol.~19, no.~12, pp. 1995--2013, 2019.

\bibitem{cont2024rough}
R.~Cont and P.~Das, ``Rough volatility: fact or artefact?'' \emph{Sankhya B}, vol.~86, no.~1, pp. 191--223, 2024.

\bibitem{matas2021simulation}
J.~Matas and J.~Posp{\'\i}{\v{s}}il, ``On simulation of rough volterra stochastic volatility models,'' \emph{arXiv preprint arXiv:2108.01999}, 2021.

\bibitem{abi2024volatility}
E.~Abi~Jaber and S.~X. Li, ``Volatility models in practice: Rough, path-dependent or markovian?'' \emph{Path-Dependent or Markovian}, 2024.

\bibitem{brandi2022multiscaling}
G.~Brandi and T.~Di~Matteo, ``Multiscaling and rough volatility: An empirical investigation,'' \emph{International Review of Financial Analysis}, vol.~84, p. 102324, 2022.

\bibitem{euch2019short}
O.~E. Euch, M.~Fukasawa, J.~Gatheral, and M.~Rosenbaum, ``Short-term at-the-money asymptotics under stochastic volatility models,'' \emph{SIAM Journal on Financial Mathematics}, vol.~10, no.~2, pp. 491--511, 2019.

\bibitem{abi2019multifactor}
E.~Abi~Jaber and O.~El~Euch, ``Multifactor approximation of rough volatility models,'' \emph{SIAM journal on financial mathematics}, vol.~10, no.~2, pp. 309--349, 2019.

\bibitem{el2018microstructural}
O.~El~Euch, M.~Fukasawa, and M.~Rosenbaum, ``The microstructural foundations of leverage effect and rough volatility,'' \emph{Finance and Stochastics}, vol.~22, pp. 241--280, 2018.

\bibitem{el2019characteristic}
O.~El~Euch and M.~Rosenbaum, ``The characteristic function of rough heston models,'' \emph{Mathematical Finance}, vol.~29, no.~1, pp. 3--38, 2019.

\bibitem{abi2019markovian}
E.~Abi~Jaber and O.~El~Euch, ``Markovian structure of the volterra heston model,'' \emph{Statistics \& Probability Letters}, vol. 149, pp. 63--72, 2019.

\bibitem{galichon2021unreasonable}
A.~Galichon, ``The unreasonable effectiveness of optimal transport in economics,'' \emph{arXiv preprint arXiv:2107.04700}, 2021.

\bibitem{torres2021survey}
L.~C. Torres, L.~M. Pereira, and M.~H. Amini, ``A survey on optimal transport for machine learning: Theory and applications,'' \emph{arXiv preprint arXiv:2106.01963}, 2021.

\bibitem{horvath2024functional}
B.~Horvath, A.~Jacquier, A.~Muguruza, and A.~S{\o}jmark, ``Functional central limit theorems for rough volatility,'' \emph{Finance and Stochastics}, pp. 1--47, 2024.

\bibitem{wiese2019deep}
M.~Wiese, L.~Bai, B.~Wood, and H.~Buehler, ``Deep hedging: learning to simulate equity option markets,'' \emph{arXiv preprint arXiv:1911.01700}, 2019.

\bibitem{ilhan2009optimal}
A.~Ilhan, M.~Jonsson, and R.~Sircar, ``Optimal static-dynamic hedges for exotic options under convex risk measures,'' \emph{Stochastic Processes and their Applications}, vol. 119, no.~10, pp. 3608--3632, 2009.

\bibitem{chakraborty2024explainable}
T.~Chakraborty, C.~Seifert, and C.~Wirth, ``Explainable bayesian optimization,'' \emph{arXiv preprint arXiv:2401.13334}, 2024.

\bibitem{islam2023comprehensive}
S.~Islam, H.~Elmekki, A.~Elsebai, J.~Bentahar, N.~Drawel, G.~Rjoub, and W.~Pedrycz, ``A comprehensive survey on applications of transformers for deep learning tasks,'' \emph{Expert Systems with Applications}, p. 122666, 2023.

\bibitem{jain2022hugging}
S.~M. Jain, ``Hugging face,'' in \emph{Introduction to Transformers for NLP: With the Hugging Face Library and Models to Solve Problems}.\hskip 1em plus 0.5em minus 0.4em\relax Springer, 2022, pp. 51--67.

\bibitem{hubinger2024sleeper}
E.~Hubinger, C.~Denison, J.~Mu, M.~Lambert, M.~Tong, M.~MacDiarmid, T.~Lanham, D.~M. Ziegler, T.~Maxwell, N.~Cheng \emph{et~al.}, ``Sleeper agents: Training deceptive llms that persist through safety training,'' \emph{arXiv preprint arXiv:2401.05566}, 2024.

\bibitem{sousa2023keep}
S.~Sousa and R.~Kern, ``How to keep text private? a systematic review of deep learning methods for privacy-preserving natural language processing,'' \emph{Artificial Intelligence Review}, vol.~56, no.~2, pp. 1427--1492, 2023.

\bibitem{nguyen2024backdoor}
T.~D. Nguyen, T.~Nguyen, P.~Le~Nguyen, H.~H. Pham, K.~D. Doan, and K.-S. Wong, ``Backdoor attacks and defenses in federated learning: Survey, challenges and future research directions,'' \emph{Engineering Applications of Artificial Intelligence}, vol. 127, p. 107166, 2024.

\bibitem{senevirathna2022survey}
T.~Senevirathna, V.~H. La, S.~Marchal, B.~Siniarski, M.~Liyanage, and S.~Wang, ``A survey on xai for beyond 5g security: technical aspects, use cases, challenges and research directions,'' \emph{arXiv preprint arXiv:2204.12822}, 2022.

\bibitem{wu2023survey}
J.~Wu, S.~Yang, R.~Zhan, Y.~Yuan, D.~F. Wong, and L.~S. Chao, ``A survey on llm-gernerated text detection: Necessity, methods, and future directions,'' \emph{arXiv preprint arXiv:2310.14724}, 2023.

\bibitem{bharati2022machine}
S.~Bharati and P.~Podder, ``Machine and deep learning for iot security and privacy: applications, challenges, and future directions,'' \emph{Security and communication networks}, vol. 2022, no.~1, p. 8951961, 2022.

\bibitem{liu2022opportunistic}
Q.~Liu, T.~Zhou, Z.~Cai, and Y.~Tang, ``Opportunistic backdoor attacks: Exploring human-imperceptible vulnerabilities on speech recognition systems,'' in \emph{Proceedings of the 30th ACM International Conference on Multimedia}, 2022, pp. 2390--2398.

\bibitem{xu2021privacy}
R.~Xu, N.~Baracaldo, and J.~Joshi, ``Privacy-preserving machine learning: Methods, challenges and directions,'' \emph{arXiv preprint arXiv:2108.04417}, 2021.

\bibitem{feng2023review}
T.~Feng, R.~Hebbar, N.~Mehlman, X.~Shi, A.~Kommineni, S.~Narayanan \emph{et~al.}, ``A review of speech-centric trustworthy machine learning: Privacy, safety, and fairness,'' \emph{APSIPA Transactions on Signal and Information Processing}, vol.~12, no.~3, 2023.

\bibitem{rigaki2023survey}
M.~Rigaki and S.~Garcia, ``A survey of privacy attacks in machine learning,'' \emph{ACM Computing Surveys}, vol.~56, no.~4, pp. 1--34, 2023.

\bibitem{hou2023large}
X.~Hou, Y.~Zhao, Y.~Liu, Z.~Yang, K.~Wang, L.~Li, X.~Luo, D.~Lo, J.~Grundy, and H.~Wang, ``Large language models for software engineering: A systematic literature review,'' \emph{arXiv preprint arXiv:2308.10620}, 2023.

\bibitem{sosnin2024certified}
P.~Sosnin, M.~N. M{\"u}ller, M.~Baader, C.~Tsay, and M.~Wicker, ``Certified robustness to data poisoning in gradient-based training,'' \emph{arXiv preprint arXiv:2406.05670}, 2024.

\bibitem{yu2024generalization}
L.~Yu, S.~Liu, Y.~Miao, X.-S. Gao, and L.~Zhang, ``Generalization bound and new algorithm for clean-label backdoor attack,'' \emph{arXiv preprint arXiv:2406.00588}, 2024.

\bibitem{yang2024comprehensive}
H.~Yang, K.~Xiang, M.~Ge, H.~Li, R.~Lu, and S.~Yu, ``A comprehensive overview of backdoor attacks in large language models within communication networks,'' \emph{IEEE Network}, 2024.

\bibitem{garrido2023bayesian}
E.~C. Garrido-Merch{\'a}n, G.~G. Piris, and M.~C. Vaca, ``Bayesian optimization of esg financial investments,'' \emph{arXiv preprint arXiv:2303.01485}, 2023.

\bibitem{frazier2018tutorial}
P.~I. Frazier, ``A tutorial on bayesian optimization,'' \emph{arXiv preprint arXiv:1807.02811}, 2018.

\bibitem{shi2023badgpt}
J.~Shi, Y.~Liu, P.~Zhou, and L.~Sun, ``Badgpt: Exploring security vulnerabilities of chatgpt via backdoor attacks to instructgpt,'' \emph{arXiv preprint arXiv:2304.12298}, 2023.

\bibitem{chen2021badpre}
K.~Chen, Y.~Meng, X.~Sun, S.~Guo, T.~Zhang, J.~Li, and C.~Fan, ``Badpre: Task-agnostic backdoor attacks to pre-trained nlp foundation models,'' \emph{arXiv preprint arXiv:2110.02467}, 2021.

\bibitem{qi2021hidden}
F.~Qi, M.~Li, Y.~Chen, Z.~Zhang, Z.~Liu, Y.~Wang, and M.~Sun, ``Hidden killer: Invisible textual backdoor attacks with syntactic trigger,'' \emph{arXiv preprint arXiv:2105.12400}, 2021.

\bibitem{wu2023attacks}
B.~Wu, Z.~Zhu, L.~Liu, Q.~Liu, Z.~He, and S.~Lyu, ``Attacks in adversarial machine learning: A systematic survey from the life-cycle perspective,'' \emph{arXiv preprint arXiv:2302.09457}, 2023.

\bibitem{zhao2024survey}
S.~Zhao, M.~Jia, Z.~Guo, L.~Gan, J.~Fu, Y.~Feng, F.~Pan, and L.~A. Tuan, ``A survey of backdoor attacks and defenses on large language models: Implications for security measures,'' \emph{arXiv preprint arXiv:2406.06852}, 2024.

\bibitem{das2024security}
B.~C. Das, M.~H. Amini, and Y.~Wu, ``Security and privacy challenges of large language models: A survey,'' \emph{arXiv preprint arXiv:2402.00888}, 2024.

\bibitem{zhao2024universal}
S.~Zhao, M.~Jia, L.~A. Tuan, F.~Pan, and J.~Wen, ``Universal vulnerabilities in large language models: Backdoor attacks for in-context learning,'' \emph{arXiv preprint arXiv:2401.05949}, 2024.

\bibitem{bruned2019rough}
Y.~Bruned, I.~Chevyrev, P.~K. Friz, and R.~Preiss, ``A rough path perspective on renormalization,'' \emph{Journal of Functional Analysis}, vol. 277, no.~11, p. 108283, 2019.

\bibitem{piczak2015esc}
K.~J. Piczak, ``Esc: Dataset for environmental sound classification,'' in \emph{Proceedings of the 23rd ACM international conference on Multimedia}, 2015, pp. 1015--1018.

\bibitem{radford2023robust}
A.~Radford, J.~W. Kim, T.~Xu, G.~Brockman, C.~McLeavey, and I.~Sutskever, ``Robust speech recognition via large-scale weak supervision,'' in \emph{International conference on machine learning}.\hskip 1em plus 0.5em minus 0.4em\relax PMLR, 2023, pp. 28\,492--28\,518.

\bibitem{barrault2023seamless}
L.~Barrault, Y.-A. Chung, M.~C. Meglioli, D.~Dale, N.~Dong, M.~Duppenthaler, P.-A. Duquenne, B.~Ellis, H.~Elsahar, J.~Haaheim \emph{et~al.}, ``Seamless: Multilingual expressive and streaming speech translation,'' \emph{arXiv preprint arXiv:2312.05187}, 2023.

\bibitem{pratap2024scaling}
V.~Pratap, A.~Tjandra, B.~Shi, P.~Tomasello, A.~Babu, S.~Kundu, A.~Elkahky, Z.~Ni, A.~Vyas, M.~Fazel-Zarandi \emph{et~al.}, ``Scaling speech technology to 1,000+ languages,'' \emph{Journal of Machine Learning Research}, vol.~25, no.~97, pp. 1--52, 2024.

\bibitem{baevski2020wav2vec}
A.~Baevski, Y.~Zhou, A.~Mohamed, and M.~Auli, ``wav2vec 2.0: A framework for self-supervised learning of speech representations,'' \emph{Advances in neural information processing systems}, vol.~33, pp. 12\,449--12\,460, 2020.

\bibitem{baevski2022data2vec}
A.~Baevski, W.-N. Hsu, Q.~Xu, A.~Babu, J.~Gu, and M.~Auli, ``Data2vec: A general framework for self-supervised learning in speech, vision and language,'' in \emph{International Conference on Machine Learning}.\hskip 1em plus 0.5em minus 0.4em\relax PMLR, 2022, pp. 1298--1312.

\bibitem{hsu2021hubert}
W.-N. Hsu, B.~Bolte, Y.-H.~H. Tsai, K.~Lakhotia, R.~Salakhutdinov, and A.~Mohamed, ``Hubert: Self-supervised speech representation learning by masked prediction of hidden units,'' \emph{IEEE/ACM transactions on audio, speech, and language processing}, vol.~29, pp. 3451--3460, 2021.

\bibitem{wu2023wav2seq}
F.~Wu, K.~Kim, S.~Watanabe, K.~J. Han, R.~McDonald, K.~Q. Weinberger, and Y.~Artzi, ``Wav2seq: Pre-training speech-to-text encoder-decoder models using pseudo languages,'' in \emph{ICASSP 2023-2023 IEEE International Conference on Acoustics, Speech and Signal Processing (ICASSP)}.\hskip 1em plus 0.5em minus 0.4em\relax IEEE, 2023, pp. 1--5.

\bibitem{koffas2022can}
S.~Koffas, J.~Xu, M.~Conti, and S.~Picek, ``Can you hear it? backdoor attacks via ultrasonic triggers,'' in \emph{Proceedings of the 2022 ACM workshop on wireless security and machine learning}, 2022, pp. 57--62.

\bibitem{shi2022audio}
C.~Shi, T.~Zhang, Z.~Li, H.~Phan, T.~Zhao, Y.~Wang, J.~Liu, B.~Yuan, and Y.~Chen, ``Audio-domain position-independent backdoor attack via unnoticeable triggers,'' in \emph{Proceedings of the 28th Annual International Conference on Mobile Computing And Networking}, 2022, pp. 583--595.

\bibitem{anwar2024foundational}
U.~Anwar, A.~Saparov, J.~Rando, D.~Paleka, M.~Turpin, P.~Hase, E.~S. Lubana, E.~Jenner, S.~Casper, O.~Sourbut \emph{et~al.}, ``Foundational challenges in assuring alignment and safety of large language models,'' \emph{arXiv preprint arXiv:2404.09932}, 2024.

\bibitem{yang2020deep}
H.~Yang, X.-Y. Liu, S.~Zhong, and A.~Walid, ``Deep reinforcement learning for automated stock trading: An ensemble strategy,'' in \emph{Proceedings of the first ACM international conference on AI in finance}, 2020, pp. 1--8.

\bibitem{kim2024financial}
A.~Kim, M.~Muhn, and V.~Nikolaev, ``Financial statement analysis with large language models,'' \emph{arXiv preprint arXiv:2407.17866}, 2024.

\bibitem{shahriari2015taking}
B.~Shahriari, K.~Swersky, Z.~Wang, R.~P. Adams, and N.~De~Freitas, ``Taking the human out of the loop: A review of bayesian optimization,'' \emph{Proceedings of the IEEE}, vol. 104, no.~1, pp. 148--175, 2015.

\bibitem{hoffmann2022training}
J.~Hoffmann, S.~Borgeaud, A.~Mensch, E.~Buchatskaya, T.~Cai, E.~Rutherford, D.~d.~L. Casas, L.~A. Hendricks, J.~Welbl, A.~Clark \emph{et~al.}, ``Training compute-optimal large language models,'' \emph{arXiv preprint arXiv:2203.15556}, 2022.

\bibitem{chen2024bathe}
Y.~Chen, H.~Li, Z.~Zheng, and Y.~Song, ``Bathe: Defense against the jailbreak attack in multimodal large language models by treating harmful instruction as backdoor trigger,'' \emph{arXiv preprint arXiv:2408.09093}, 2024.

\bibitem{cheng2024transferring}
P.~Cheng, Z.~Wu, T.~Ju, W.~Du, and Z.~Z.~G. Liu, ``Transferring backdoors between large language models by knowledge distillation,'' \emph{arXiv preprint arXiv:2408.09878}, 2024.

\bibitem{zhang2024badmerging}
J.~Zhang, J.~Chi, Z.~Li, K.~Cai, Y.~Zhang, and Y.~Tian, ``Badmerging: Backdoor attacks against model merging,'' \emph{arXiv preprint arXiv:2408.07362}, 2024.

\bibitem{yu2024shadow}
J.~Yu, Y.~Yu, X.~Wang, Y.~Lin, M.~Yang, Y.~Qiao, and F.-Y. Wang, ``The shadow of fraud: The emerging danger of ai-powered social engineering and its possible cure,'' \emph{arXiv preprint arXiv:2407.15912}, 2024.

\bibitem{zhou2024comparison}
X.~Zhou, D.-M. Tran, T.~Le-Cong, T.~Zhang, I.~C. Irsan, J.~Sumarlin, B.~Le, and D.~Lo, ``Comparison of static application security testing tools and large language models for repo-level vulnerability detection,'' \emph{arXiv preprint arXiv:2407.16235}, 2024.

\bibitem{liu2024imposter}
X.~Liu, L.~Li, T.~Xiang, F.~Ye, L.~Wei, W.~Li, and N.~Garcia, ``Imposter. ai: Adversarial attacks with hidden intentions towards aligned large language models,'' \emph{arXiv preprint arXiv:2407.15399}, 2024.

\bibitem{sheshadri2024targeted}
A.~Sheshadri, A.~Ewart, P.~Guo, A.~Lynch, C.~Wu, V.~Hebbar, H.~Sleight, A.~C. Stickland, E.~Perez, D.~Hadfield-Menell \emph{et~al.}, ``Targeted latent adversarial training improves robustness to persistent harmful behaviors in llms,'' \emph{arXiv preprint arXiv:2407.15549}, 2024.

\bibitem{verma2024operationalizing}
A.~Verma, S.~Krishna, S.~Gehrmann, M.~Seshadri, A.~Pradhan, T.~Ault, L.~Barrett, D.~Rabinowitz, J.~Doucette, and N.~Phan, ``Operationalizing a threat model for red-teaming large language models (llms),'' \emph{arXiv preprint arXiv:2407.14937}, 2024.

\bibitem{achintalwar2024detectors}
S.~Achintalwar, A.~A. Garcia, A.~Anaby-Tavor, I.~Baldini, S.~E. Berger, B.~Bhattacharjee, D.~Bouneffouf, S.~Chaudhury, P.-Y. Chen, L.~Chiazor \emph{et~al.}, ``Detectors for safe and reliable llms: Implementations, uses, and limitations,'' \emph{arXiv preprint arXiv:2403.06009}, 2024.

\bibitem{berrones2010bayesian}
A.~Berrones, ``Bayesian inference based on stationary fokker-planck sampling,'' \emph{Neural Computation}, vol.~22, no.~6, pp. 1573--1596, 2010.

\bibitem{marion2024implicit}
P.~Marion, A.~Korba, P.~Bartlett, M.~Blondel, V.~De~Bortoli, A.~Doucet, F.~Llinares-L{\'o}pez, C.~Paquette, and Q.~Berthet, ``Implicit diffusion: Efficient optimization through stochastic sampling,'' \emph{arXiv preprint arXiv:2402.05468}, 2024.

\bibitem{melnik2024video}
A.~Melnik, M.~Ljubljanac, C.~Lu, Q.~Yan, W.~Ren, and H.~Ritter, ``Video diffusion models: A survey,'' \emph{arXiv preprint arXiv:2405.03150}, 2024.

\bibitem{xing2023survey}
Z.~Xing, Q.~Feng, H.~Chen, Q.~Dai, H.~Hu, H.~Xu, Z.~Wu, and Y.-G. Jiang, ``A survey on video diffusion models,'' \emph{arXiv preprint arXiv:2310.10647}, 2023.

\bibitem{zhang2023text}
C.~Zhang, C.~Zhang, M.~Zhang, and I.~S. Kweon, ``Text-to-image diffusion models in generative ai: A survey,'' \emph{arXiv preprint arXiv:2303.07909}, 2023.

\bibitem{cao2024survey}
H.~Cao, C.~Tan, Z.~Gao, Y.~Xu, G.~Chen, P.-A. Heng, and S.~Z. Li, ``A survey on generative diffusion models,'' \emph{IEEE Transactions on Knowledge and Data Engineering}, 2024.

\bibitem{yang2023diffusion}
L.~Yang, Z.~Zhang, Y.~Song, S.~Hong, R.~Xu, Y.~Zhao, W.~Zhang, B.~Cui, and M.-H. Yang, ``Diffusion models: A comprehensive survey of methods and applications,'' \emph{ACM Computing Surveys}, vol.~56, no.~4, pp. 1--39, 2023.

\bibitem{hu2024large}
Y.~Hu, C.~Chen, C.-H.~H. Yang, R.~Li, C.~Zhang, P.-Y. Chen, and E.~Chng, ``Large language models are efficient learners of noise-robust speech recognition,'' \emph{arXiv preprint arXiv:2401.10446}, 2024.

\bibitem{rebonato2002theory}
R.~Rebonato, ``Theory and practice of model risk management,'' \emph{Modern Risk Management: A History’, RiskWaters Group, London}, pp. 223--248, 2002.

\bibitem{fliess2010delta}
M.~Fliess and C.~Join, ``Delta hedging in financial engineering: Towards a model-free approach,'' in \emph{18th Mediterranean Conference on Control and Automation, MED'10}.\hskip 1em plus 0.5em minus 0.4em\relax IEEE, 2010, pp. 1429--1434.

\bibitem{fliess2012preliminary}
------, ``Preliminary remarks on option pricing and dynamic hedging,'' in \emph{2012 1st International Conference on Systems and Computer Science (ICSCS)}.\hskip 1em plus 0.5em minus 0.4em\relax IEEE, 2012, pp. 1--6.

\bibitem{choe2016girsanov}
G.~H. Choe, ``Girsanov’s theorem,'' in \emph{Stochastic Analysis for Finance with Simulations}.\hskip 1em plus 0.5em minus 0.4em\relax Springer, 2016, pp. 137--145.

\bibitem{bondi2024rough}
A.~Bondi, S.~Pulido, and S.~Scotti, ``The rough hawkes heston stochastic volatility model,'' \emph{Mathematical Finance}, 2024.

\bibitem{horvath2020volatility}
B.~Horvath, A.~Jacquier, and P.~Tankov, ``Volatility options in rough volatility models,'' \emph{SIAM Journal on Financial Mathematics}, vol.~11, no.~2, pp. 437--469, 2020.

\bibitem{mccrickerd2019spatially}
R.~McCrickerd, ``On spatially irregular ordinary differential equations and a pathwise volatility modelling framework,'' \emph{arXiv preprint arXiv:1902.01673}, 2019.

\bibitem{keller2018affine}
M.~Keller-Ressel, M.~Larsson, and S.~Pulido, ``Affine rough models,'' \emph{arXiv preprint arXiv:1812.08486}, 2018.

\bibitem{sakurai2024learning}
R.~Sakurai, H.~Takahashi, and K.~Miyamoto, ``Learning tensor networks with parameter dependence for fourier-based option pricing,'' \emph{arXiv preprint arXiv:2405.00701}, 2024.

\bibitem{zhang2013fast}
S.~Zhang and L.~Wang, ``Fast fourier transform option pricing with stochastic interest rate, stochastic volatility and double jumps,'' \emph{Applied Mathematics and Computation}, vol. 219, no.~23, pp. 10\,928--10\,933, 2013.

\bibitem{wong2011fft}
H.~Y. Wong and P.~Guan, ``An fft-network for l{\'e}vy option pricing,'' \emph{Journal of Banking \& Finance}, vol.~35, no.~4, pp. 988--999, 2011.

\bibitem{wang2022pricing}
W.~Wang and X.~Hu, ``Pricing israeli option with time-changed compensation by an fft-based high-order multinomial tree in l{\'e}vy markets,'' \emph{Computational Intelligence and Neuroscience}, vol. 2022, no.~1, p. 9682292, 2022.

\bibitem{hurd2010fourier}
T.~R. Hurd and Z.~Zhou, ``A fourier transform method for spread option pricing,'' \emph{SIAM Journal on Financial Mathematics}, vol.~1, no.~1, pp. 142--157, 2010.

\bibitem{abi2024signature}
E.~Abi~Jaber and L.-A. G{\'e}rard, ``Signature volatility models: pricing and hedging with fourier,'' \emph{Available at SSRN}, 2024.

\bibitem{aurzada2020ruin}
F.~Aurzada and M.~Buck, ``Ruin probabilities in the cram{\'e}r--lundberg model with temporarily negative capital,'' \emph{European Actuarial Journal}, vol.~10, no.~1, pp. 261--269, 2020.

\bibitem{braunsteins2023cramer}
P.~Braunsteins and M.~Mandjes, ``The cram{\'e}r-lundberg model with a fluctuating number of clients,'' \emph{Insurance: Mathematics and Economics}, vol. 112, pp. 1--22, 2023.

\bibitem{xiong2011ruin}
S.~Xiong and W.-S. Yang, ``Ruin probability in the cram{\'e}r--lundberg model with risky investments,'' \emph{Stochastic processes and their applications}, vol. 121, no.~5, pp. 1125--1137, 2011.

\bibitem{weber2024critical}
M.~J. Weber, ``Critical probabilistic characteristics of the cram{\'e}r model for primes and arithmetical properties,'' \emph{Indian Journal of Pure and Applied Mathematics}, pp. 1--27, 2024.

\bibitem{chen2023geometric}
D.~Chen, Z.~Zhou, J.-P. Mei, C.~Shen, C.~Chen, and C.~Wang, ``A geometric perspective on diffusion models,'' \emph{arXiv preprint arXiv:2305.19947}, 2023.

\bibitem{yang2024neural}
H.~Yang, A.~Hasan, Y.~Ng, and V.~Tarokh, ``Neural mckean-vlasov processes: Distributional dependence in diffusion processes,'' in \emph{International Conference on Artificial Intelligence and Statistics}.\hskip 1em plus 0.5em minus 0.4em\relax PMLR, 2024, pp. 262--270.

\bibitem{coghi2018pathwise}
M.~Coghi, J.-D. Deuschel, P.~Friz, and M.~Maurelli, ``Pathwise mckean-vlasov theory with additive noise,'' \emph{arXiv preprint arXiv:1812.11773}, 2018.

\bibitem{carmona2013control}
R.~Carmona, F.~Delarue, and A.~Lachapelle, ``Control of mckean--vlasov dynamics versus mean field games,'' \emph{Mathematics and Financial Economics}, vol.~7, pp. 131--166, 2013.

\bibitem{han2024learning}
J.~Han, R.~Hu, and J.~Long, ``Learning high-dimensional mckean--vlasov forward-backward stochastic differential equations with general distribution dependence,'' \emph{SIAM Journal on Numerical Analysis}, vol.~62, no.~1, pp. 1--24, 2024.

\bibitem{qian2022mckean}
Z.~Qian and Y.~Yao, ``Mckean--vlasov type stochastic differential equations arising from the random vortex method,'' \emph{Partial Differential Equations and Applications}, vol.~3, no.~1, p.~7, 2022.

\bibitem{adams2022large}
D.~Adams, G.~Dos~Reis, R.~Ravaille, W.~Salkeld, and J.~Tugaut, ``Large deviations and exit-times for reflected mckean--vlasov equations with self-stabilising terms and superlinear drifts,'' \emph{Stochastic Processes and their Applications}, vol. 146, pp. 264--310, 2022.

\bibitem{meherrem2019maximum}
S.~Meherrem and M.~Hafayed, ``Maximum principle for optimal control of mckean-vlasov fbsdes with l{\'e}vy process via the differentiability with respect to probability law,'' \emph{Optimal Control Applications and Methods}, vol.~40, no.~3, pp. 499--516, 2019.

\end{thebibliography}

\end{document}